\documentclass{article}

  \usepackage[preprint]{neurips_2025}

\usepackage[utf8]{inputenc} 
\usepackage[T1]{fontenc}    
\usepackage{hyperref}       
\usepackage{url}            
\usepackage{booktabs}       
\usepackage{amsfonts}       
\usepackage{nicefrac}       
\usepackage{microtype}      
\usepackage{xcolor}         

\usepackage{xspace}
\usepackage{multirow}
\usepackage{colortbl}
\usepackage{float}
\usepackage{enumitem}
\usepackage{amsmath}
\usepackage{hhline}
\usepackage{graphicx}
\usepackage{tikz}

\newcommand{\circnum}[1]{%
  \tikz[baseline=(char.base)]{
    \node[shape=circle, fill=black, text=white, inner sep=1pt] (char) {\scriptsize #1};
  }%
}

\newcommand{\para}[1]{\noindent \textbf{#1.}}
\newcommand{\papername}{SemanticDialect\xspace}
\newcommand{\formatname}{FB4\xspace}

\title{SemanticDialect: Semantic-Aware Mixed-Format Quantization for Video Diffusion Transformers}

\author{%
  Wonsuk Jang \\
  Stanford University \\
  \texttt{wsjang@stanford.edu} \\
  \And
  Thierry Tambe \\
  Stanford University \\
  \texttt{ttambe@stanford.edu} \\
}

\begin{document}

\maketitle


\begin{abstract}
Diffusion Transformers (DiT) achieve strong video generation quality, but their memory and compute costs hinder edge deployment. Quantization can reduce these costs, yet existing methods often degrade video quality under high activation variation and the need to preserve semantic/temporal coherence. We propose \textbf{\papername}, which advances recent block-wise mixed-format quantization---selecting a per-block optimal format (a \emph{dialect}) from multiple candidates (a \emph{formatbook})---by scaling the formatbook with lookup tables for quantization error and quantized values, enabling efficient per-block format selection and quantization at low online cost. We also introduce activation decomposition that reduces quantization error by re-quantizing and adding back residual errors, with attention-guided salient token selection. We further propose semantic-aware dialect assignment (SeDA) to improve quantized value consistency by sharing a sub-formatbook among semantically correlated tokens. Experiments on video DiT (VDiT) models show that \papername outperforms prior VDiT quantization methods and fine-grained block-wise format baselines, while approaching FP16 quality on Open-Sora~2.0. 

\end{abstract}
\begin{figure}[H]
  \centering
  \includegraphics[page=11, width=0.8\columnwidth]{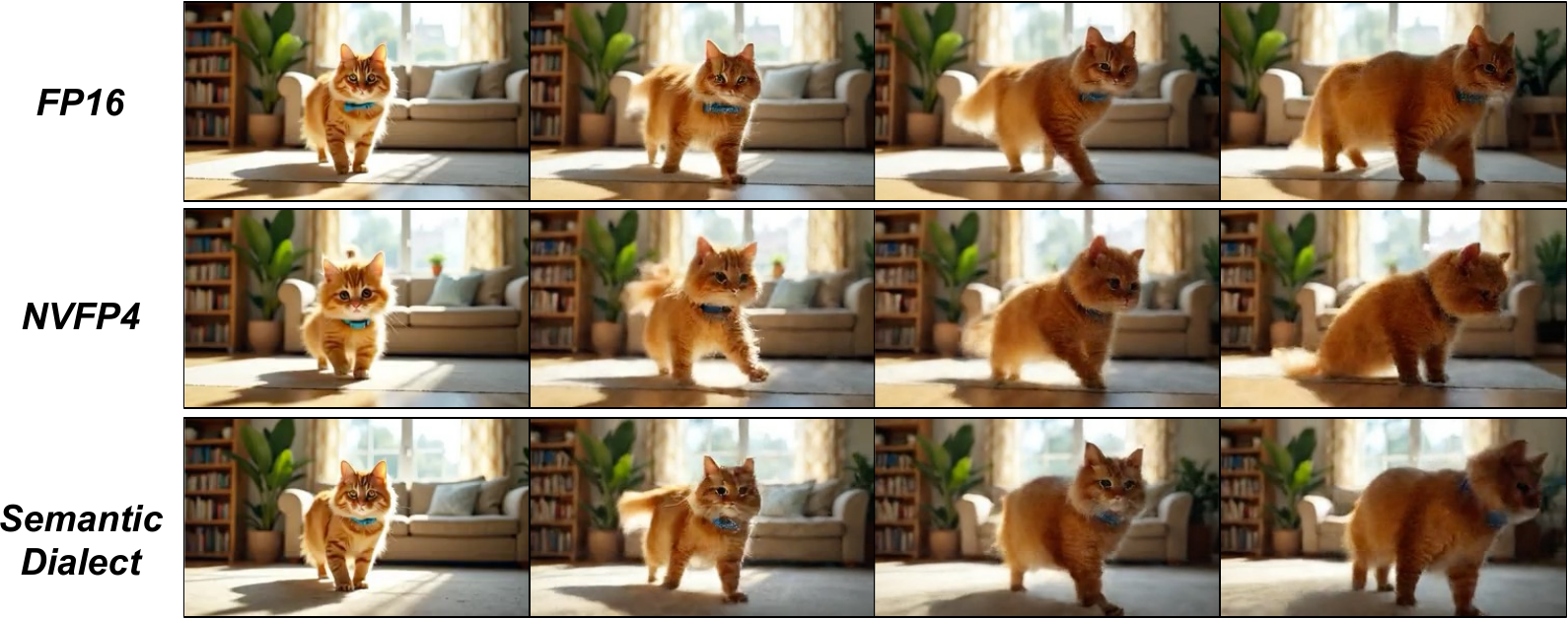}
  \caption{\papername is a post-training quantization method for video diffusion transformers. It quantizes Open-Sora~2.0 to 4-bit weights and activations while maintaining near-FP16 visual quality.}
  \label{fig:first_fig}
\end{figure}
\section{Introduction}
\label{sec:intro}


Diffusion Transformers (DiTs)~\citep{peebles2023scalable} have gained prominence in video generation~\citep{ma2024latte,yang2024cogvideox,peng2025open2}, as their scalability and capacity to capture long-range spatiotemporal context, and are increasingly deployed as a backbone of many applications such as Physical AI~\citep{ye2026world}. However, their large parameter counts and necessity of iterative denoising (often tens of steps) substantially increase compute and memory demands, exacerbated by the long sequence lengths of multi-frame video---making edge deployment challenging.


Quantization mitigates these costs by reducing memory footprints and data movement, while enabling efficient low-bit arithmetic for enhanced throughput and energy efficiency~\citep{kim2023squeezellm,xiao2023smoothquant,darvish2023shared,cao2024abs}. However, prior DiT quantization methods~\citep{wu2024ptq4dit,chen2025q} often struggle to preserve text-to-video generation quality~\citep{zhao2024vidit}. This degradation stems from two key challenges: (i) high activation variation across tokens and timesteps, where a few large-magnitude outliers dominate the scaling factor, reducing effective quantization precision for the majority of elements~\citep{liu2023llm}, and (ii) strong spatiotemporal correlations across frames, which standard MSE-based quantization objectives often fail to capture.

In response to the first challenge, fine-grained block-wise quantization has become widely adopted~\citep{dai2021vs,darvish2023shared,hu2026m,lo2024nanoscaling}, assigning separate scaling factors to small blocks to localize outlier impact. This trend is reflected in industry efforts such as the OCP Microscaling (MX) specification, which standardizes block-wise formats using power-of-two scales~\citep{ocp_mx_v1}. Recent accelerators from NVIDIA and AMD have also embraced MX formats~\citep{nvidia_nvfp4_blog,amd_cdna4_wp}. To further improve accuracy, higher-precision block-scale variants (e.g., NVFP4) have been introduced, albeit at the expense of increased metadata and runtime overhead for scale normalization and rescaling. Despite these advances, a single low-precision format still struggles to match the diverse per-block value distributions.

Motivated by this, prior work proposes a mixed-format quantization that selects an optimal per-block format from a predefined set (the \emph{formatbook}) to better match block value distributions~\citep{jangblockdialect}.  
However, extending this strategy to VDiTs is non-trivial for three reasons: (i) VDiT activations exhibit higher variability, calling for a larger, more adaptive formatbook; (ii) on-the-fly format selection and quantization impose substantial overhead, as the complexity of decision logic grows with the number of representable values across all candidate formats; and (iii) independent block quantization ignores the semantic and temporal correlations essential for video consistency.

To tackle these challenges, we propose \textbf{\papername}, a VDiT post-training quantization (PTQ) framework built on the \textbf{\formatname} format, which enables calibration-free block-wise mixed-format 4-bit quantization. To handle high activation variability, \formatname employs a 32-entry formatbook with lookup table (LUT)-based online format selection and quantization. To mitigate quality loss in quantization-sensitive layers, we introduce attention-guided \emph{activation decomposition}, which reduces quantization error by re-quantizing the residual and adding it back, without conventional mixed-precision overhead. Finally, to avoid block-wise over-specialization---where inconsistent quantization of identical activation values across blocks undermines spatiotemporal consistency---we propose \emph{semantic-aware dialect assignment (SeDA)}, which encourages semantically aligned tokens to share a subset of the formatbook. Our contributions are as follows:

\begin{itemize}
    \item We introduce \formatname, a 4-bit number format enabling calibration-free, online mixed-format quantization via per-block LUT-based format selection from a 32-entry formatbook.
    \item We propose \papername, a VDiT quantization framework that integrates \formatname with activation decomposition for sensitive layers and SeDA for spatiotemporal consistency.
   \item We demonstrate that \papername outperforms existing VDiT quantization methods and block-wise format baselines on Open-Sora 1.0 and 2.0, two distinct VDiT architectures.
\end{itemize}

\section{Related Work}
\label{sec:related_work}

\subsection{Video Diffusion Transformers (VDiT)} 
DiTs~\citep{peebles2023scalable} replace traditional convolutional U-Net backbones with Transformer blocks, serving as a scalable foundation for modern video generation. Similar to prior diffusion models, DiTs iteratively denoise latents over multiple timesteps, with Transformers better capturing global context over large token sequences. VDiTs commonly exhibit two architectural patterns: (i) injecting external conditions (e.g., text and timestep) via cross-attention or adaptive LayerNorm (adaLN) modulation, and (ii) spatiotemporal modeling via either factorized attention that decouples spatial and temporal interactions~\citep{ma2024latte, zheng2024open} or full 3D attention~\citep{peng2025open2, yang2024cogvideox,kong2024hunyuanvideo} that jointly attends across space–time tokens paired with 3D rotary positional embeddings (3D RoPE). 
To enhance text-to-video alignment, many VDiT models use classifier-free guidance (CFG), combining the outputs of a conditional branch (with condition tokens) and an unconditional branch (with null tokens).

\subsection{VDiT Quantization}
A growing body of work explores post-training quantization for DiTs. Q-DiT and PTQ4DiT~\citep{chen2025q, wu2024ptq4dit} aim to stabilize activation quantization across channels and timesteps, while SVDQuant~\citep{li2024svdquant} reduces error by incorporating a lightweight SVD-based low-rank compensation branch. However, these designs do not fully address the pronounced spatiotemporal variation in VDiTs, nor do they optimize for video-specific quality metrics beyond MSE. Recent VDiT-focused approaches improve quality through video-aware heuristics, including metric-aware mixed precision (ViDiT-Q~\citep{zhao2024vidit}), inter-frame distribution distillation to better preserve temporal behavior (Q-VDiT~\citep{feng2025q}), and specialized calibration-data selection (S$^2$Q-VDiT~\citep{ffeng2025s}). In contrast, \papername targets the root cause---diverse block-level distributions across layers and timesteps---through online, per-block format selection to match local statistics without intensive calibration. We also introduce uniform-precision activation decomposition for quantization-sensitive layers without mixed-precision activations, and semantic-aware dialect assignment (SeDA) to preserve spatiotemporal consistency of quantized values.

\subsection{Fine-Grained Block-Wise Mixed-Format Quantization}
Fine-grained block-wise quantization has become widely adopted~\citep{dai2021vs,darvish2023shared,hu2026m,lo2024nanoscaling,rouhani2023microscaling} by assigning separate scaling factors to small blocks, effectively localizing outlier impact. However, mapping normalized values to a fixed low-precision format (e.g., FP4) can still incur large error due to limited representational capacity. Mixed-format methods attempt to improve value representation by selecting among multiple candidates: RaZeR~\citep{chen2025razer} repurposes redundant $\pm0$ encodings in NVFP4 into predefined special values, and BlockDialect~\citep{jangblockdialect} selects a per-block format from a predefined set (a \emph{formatbook}) while retaining power-of-two scaling. However, their expressiveness is often limited by predefined special values or small formatbooks, and scaling to a larger set of candidates is challenging because they require costly repeated quantization or comparison logic over individual representable values.
We address this by scaling to a larger, more expressive formatbook with low online overhead via LUT-based format selection and quantization. We use the term \emph{dialect} to refer to each same-precision candidate with slightly different representable values in the formatbook.

\begin{figure}[t]
  \centering
  \includegraphics[page=1, width=0.9\columnwidth]{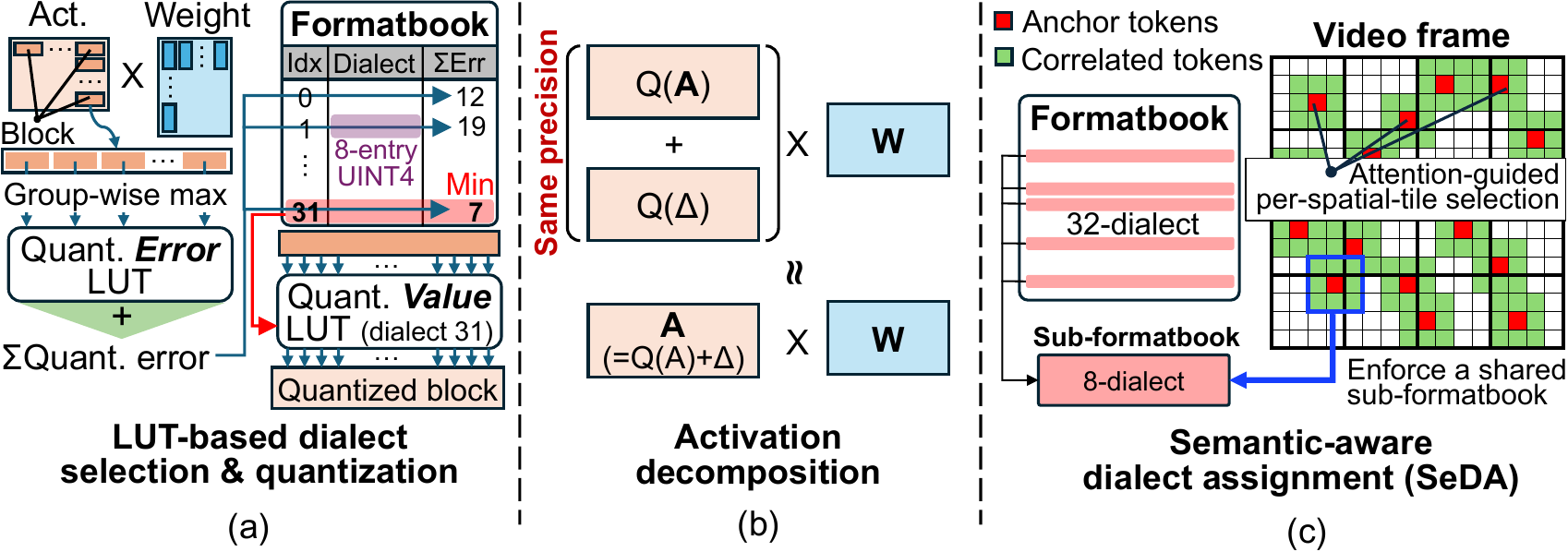}
  \caption{SemanticDialect overview. (a) LUT-based format selection and quantization. (b) Activation decomposition for quantization-sensitive layers. (c) SeDA for consistent block-wise quantization.}
  \label{fig:overview}
\end{figure}
\section{SemanticDialect: Semantic-Aware Mixed-Format Quantization for Video Diffusion Transformers}
\label{sec:semantic_dialect}

\begin{figure}[t]
  \centering
  \includegraphics[page=2, width=0.95\columnwidth]{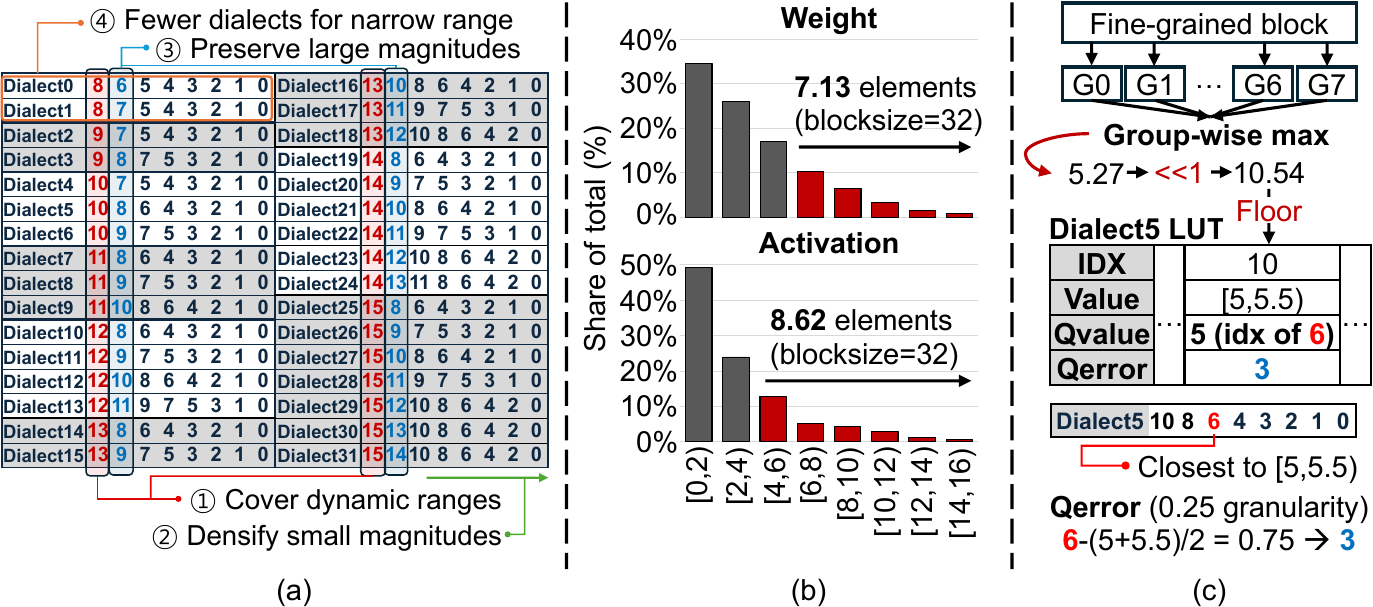}
\caption{Proposed formatbook and lookup table. (a) Four formatbook construction ground rules. \linebreak (b) Block-wise normalized value distributions. (c) Lookup table details and data-to-index mapping.}
  \label{fig:lut}
\end{figure}
Figure~\ref{fig:overview} summarizes our approach. We address four challenges to enable quality-preserving, fine-grained block-wise mixed-format quantization for VDiTs: (1) designing an expressive formatbook; (2) enabling efficient online per-block dialect selection and quantization; (3) mitigating errors in quantization-sensitive layers; and (4) ensuring semantic awareness across spatial and temporal axes.
\subsection{Constructing a Highly Expressive Formatbook}
\label{sec:semantic_dialect:formatbook}

\para{Block Normalization and 4-bit Representation}
For each block, we compute a shared power-of-two scale by  applying $\lfloor \log_2(\cdot)\rfloor$ to the block's maximum magnitude. We then normalize each block so that its values lie in $[0,2)$, with the maximum in $[1,2)$. Elements are represented by a 1-bit sign and a 3-bit index (i.e., 4-bit quantization) that maps to one of eight representable magnitudes in the dialect. These magnitudes are selected from 4-bit unsigned integers (0-15), enabling hardware-efficient low-precision multiply--accumulate (MAC). For seamless implementation, we offset the shared exponent by $3$ (equivalently, scale the normalized values by $2^3$), mapping $[0,2)$ to $[0,16)$ (with the maximum in $[8,16)$) so magnitudes can be directly encoded as 4-bit unsigned integers.

\para{Formatbook Design}
Since the 16-dialect formatbook in prior work~\citep{jangblockdialect} cannot fully capture the higher variation in VDiT weights and activations (see Table~\ref{tab:main}), we build a larger \emph{32-dialect} formatbook (Fig.~\ref{fig:lut}a) based on four design rules. (1) \emph{Cover all dynamic ranges:} since power-of-two scaling discretizes range, insufficient coverage can waste dynamic range or leave values underrepresented~\citep{jangblockdialect, lo2024nanoscaling}. (2) \emph{Densify small magnitudes:} Open-Sora~2.0 profiling (Fig.~\ref{fig:lut}b) shows most values cluster near zero, so we allocate more points to small magnitudes. (3) \emph{Preserve large magnitudes:} 
large-magnitude values often dominate MAC outputs yet suffer high quantization error under small-magnitude–biased dialects due to sparse high-range coverage~\citep{cook2025four}; thus, each dynamic range includes dialects that better represent large magnitudes. (4) \emph{Use fewer dialects for narrower ranges:} this reduces dialect ID (DID) metadata and simplifies per-block dialect selection. The total per-block metadata overhead is 10 bits (5-bit DID and 5-bit shared exponent). Although this rule-based formatbook may not be globally optimal, our goal is to demonstrate scalable large-formatbook mixed-format quantization rather than exhaustively optimizing the formatbook. Appendix~\ref{app:format_usage} confirms that all dialects are meaningfully leveraged.

\subsection{LUT-Based Per-Block Dialect Selection and Quantization}
\label{sec:semantic_dialect:lut}
\para{Group-wise Maximum Extraction}
A straightforward dialect selection method computes the MSE for each of the 32 dialects across all block elements (e.g., 32-element) and picks the minimum. However, this full comparison is costly across numerous blocks. Profiling (Fig.~\ref{fig:lut}b) on Open-Sora~2.0 shows that only a small fraction of elements have large magnitudes, yet they disproportionately contribute to MAC outputs. To balance accuracy and compute overhead, we restrict MSE evaluation to the top-$k$ largest-magnitude values (e.g., $k=8$). Since exact top-$k$ requires sorting, we instead approximate by partitioning each block into $k$ groups (\emph{num\_groups $=k$}) and taking the maximum from each group (Fig.~\ref{fig:lut}c), enabling parallel group-wise computation.

\para{LUT-Based Quantization Error Approximation}
Even with group-wise maximum extraction, evaluating quantization MSE remains compute-intensive. This is because our dialects span diverse distributions, so quantization cannot be reduced to a simple shift-and-round operation as in standard 4-bit formats (e.g., INT4, FP4). Instead, for each dialect we must (i) find the nearest representable value and (ii) accumulate its deviation from the input. To make this efficient for a large formatbook, we introduce two lookup tables per dialect, \texttt{Qvalue} and \texttt{Qerror} (Fig.~\ref{fig:lut}c): \texttt{Qvalue} returns the 3-bit index of the nearest representable value, while \texttt{Qerror} provides an approximate quantization error.

Since non-integer inputs (e.g., FP16) cannot directly index a LUT, we apply a shift-and-truncate to map them to integer bin indices. Because representable values are integers (with decision boundaries at half-integers), \texttt{Qvalue} uses 0.5-wide bins so that each bin maps to the \emph{exact} quantized index. 
For \texttt{Qerror}, we \emph{approximate} absolute quantization error using the midpoint error of each 0.5-wide bin, rather than squared error, which can amplify midpoint estimation bias. Dialect selection then reduces to summing \texttt{Qerror} over group-wise maxima and choosing the dialect with minimum approximate error. Once selected, quantizing all elements proceeds efficiently via direct \texttt{Qvalue} queries.

\para{Two-stage Dialect Selection}
A remaining challenge is the cost of comparing all 32 dialects to select the best per-block dialect. Since matching the block's dynamic range is critical to avoid underestimating or wasting range, we adopt a two-stage approach similar to the approach in~\citep{jangblockdialect}. In the first stage, we select a sub-formatbook based on the block maximum. In the second stage, we compute the approximate error and compare only the dialects within that selected subset. We refer to our 4-bit quantization format, which utilizes the formatbook with LUT-based dialect selection and quantization, as \textbf{\formatname} (FormatBook 4-bit format) for the remainder of the paper.

\subsection{Compensating Quantization-Sensitive Layers}
\label{sec:semantic_dialect:decomp}
\para{Layer-wise Profiling Results}
\papername primarily targets linear layers (\texttt{nn.Linear}), which perform activation--weight MACs and dominate inference latency~\citep{zhao2024vidit}. For each layer, we profile (i) \emph{sensitivity} by quantizing only that layer in an otherwise FP16 model, and (ii) \emph{recovery potential} by keeping only that layer in FP16 in an otherwise \formatname-quantized model. Appendix~\ref{app:sensitivity} shows that \emph{modulation layers} are the most quantization-sensitive in Open-Sora~2.0 (Flux-based~\citep{labs2025flux}), likely because their inputs---single vectors encoding text and timestep---are already highly compressed. In Open-Sora~1.0 (STDiT-based~\citep{zheng2024open}), these layers are relatively small and typically kept in FP16~\citep{zhao2024vidit,feng2025q}. Beyond modulation, we find the \emph{final MLP linear layer}---often sensitive since it directly precedes the layer output~\citep{yan2026d,ashkboos2024quik}---and the \emph{temporal attention QKV projection} to also be highly sensitive.

\begin{figure}[t]
  \centering
  \includegraphics[page=3, width=0.75\columnwidth]{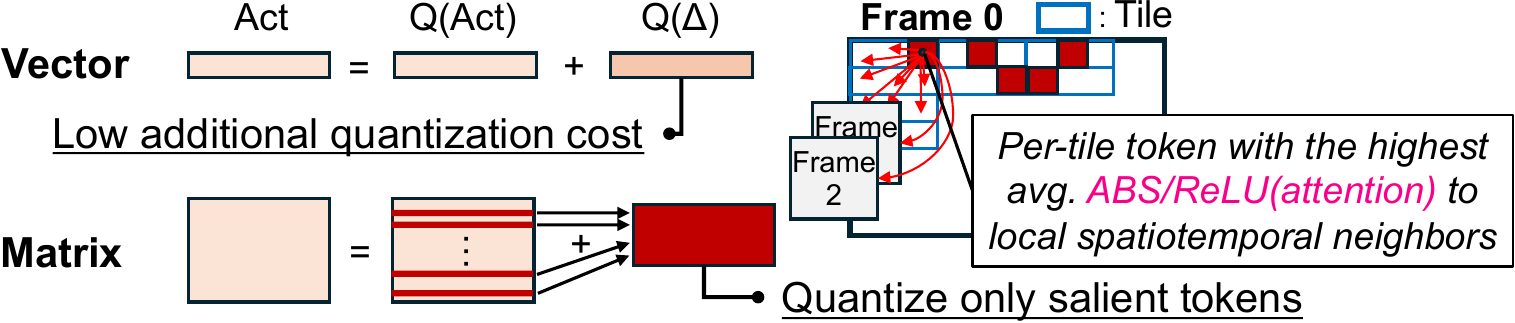}
\caption{Activation decomposition for vector and matrix activations.}
  \label{fig:act_decomp}
\end{figure}
\para{Activation Decomposition}
In diffusion transformers, activations are often harder to quantize than weights due to timestep-dependent, channel-wise outliers~\citep{ding2025post, chen2025q}. In contrast, weights are timestep-invariant and can be quantized offline, allowing exhaustive dialect selection without runtime approximations. We therefore focus on improving \emph{activation} quantization in sensitive layers. Mixed precision (i.e., using higher precision for sensitive layers) is a straightforward option, but it is undesirable: it complicates GPU kernels (multiple data types and conversions, fewer fusion opportunities), increases memory-management complexity, and requires extra datapaths or format-specific kernels---especially for non-standard formats such as \formatname or on dedicated hardware.

Instead, we propose \emph{activation decomposition}. As shown in Figure~\ref{fig:act_decomp}, we quantize the activation, forming residual $\Delta$ where $Act = Q(Act) + \Delta$, then re-quantize $\Delta$ to obtain $Q(Act)+Q(\Delta)$ as a closer approximation to $Act$ using the same low-precision format. Since decomposition applies only to activations, the parameter count remains unchanged and a linear layer is approximated as $Act\cdot W+b \approx \bigl(Q(Act)+Q(\Delta)\bigr)W + b$. The modulation layer in Open-Sora~2.0 is particularly suitable because its activation is a single \emph{vector}, incurring negligible re-quantization overhead.

For \emph{matrix} activations, decomposing all tokens is costly, substantially increasing the effective bitwidth (two 4-bit values per token, $\approx 8$ bits plus metadata). We therefore apply decomposition only to one salient token per tile, where a tile is a local group of tokens. Concretely, we score candidates by their mean pre-softmax attention score ($Q\cdot K^{\mathsf T}$) with respect to a local spatiotemporal neighborhood and select the highest-scoring token in each tile.
To prevent large negative scores from canceling meaningful correlations, we use ReLU for temporal attention to focus on positive attention connections across frames, while using ABS for spatial / 3D attention since both positive and negative responses can be informative (similarity and contrast). Furthermore, in some layers, allocating the salient token budget exclusively to the \emph{conditional} CFG branch proves more effective than splitting it evenly across conditional and unconditional branches. Detailed evaluation is provided in Appendix~\ref{app:salient_token}.

\subsection{Incorporating Semantic Awareness Across Spatial and Temporal Axes}
\label{sec:semantic_dialect:sefa}

\para{Block-Wise Over-Specialization} Although block-wise mixed-format quantization better matches local distributions, a large formatbook can introduce dialect inconsistency: the same value may be assigned different dialects across blocks due to minor shifts in block statistics. For example, the same spatial token across frames, or spatially adjacent tokens within a semantic region may be quantized inconsistently, potentially leading to visual artifacts. 
In short, fine-grained dialect matching can over-specialize to local blocks---optimizing the ``trees'' while harming the ``forest.''

\begin{figure}[t]
  \centering
  \includegraphics[page=4, width=0.85\columnwidth]{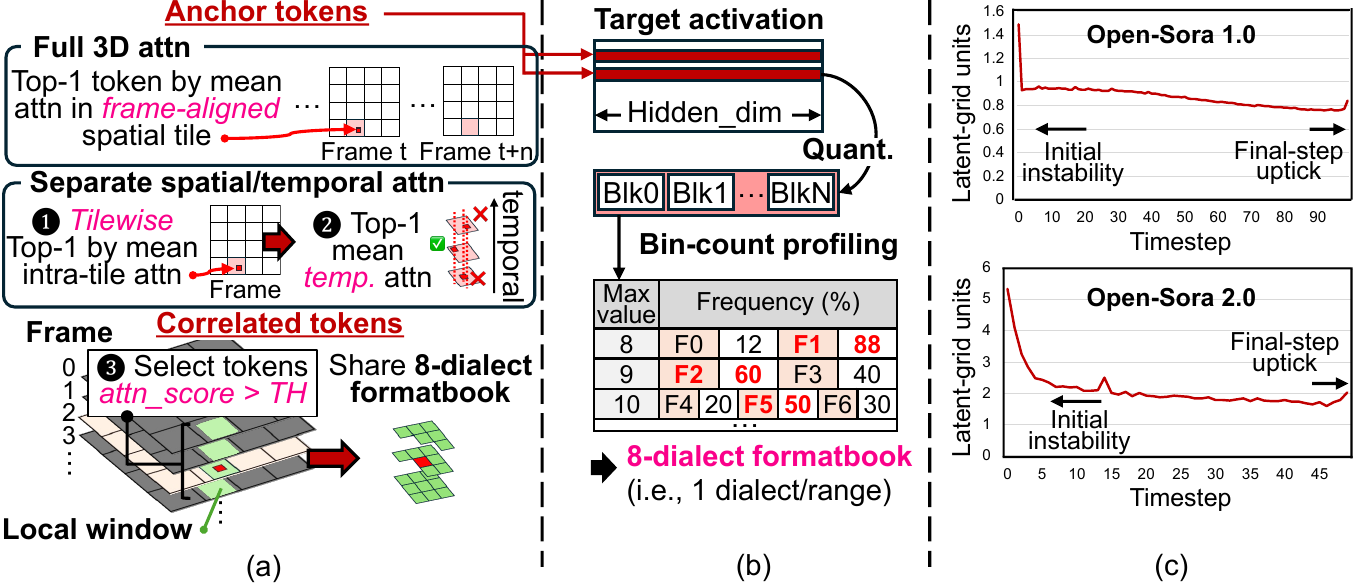}
\caption{Semantic-Aware Dialect Assignment (SeDA). (a) Anchor/correlated token selection. (b) Sub-formatbook construction. (c) Average anchor-token displacement across denoising timesteps.}

  \label{fig:seda}
\end{figure}
\para{Semantic-Aware Dialect Assignment (SeDA)}
To mitigate this over-specialization, we encourage semantically related tokens to use consistent dialect assignments. However, forcing a single shared dialect can degrade video quality because their blocks may have different dynamic ranges. Instead, we enforce consistency at the \emph{sub-formatbook} level: related tokens share the same 8-dialect sub-formatbook, which provides one dialect per dynamic range (mapped to block maxima 8--15).

\para{Constructing an 8-Dialect Sub-Formatbook}
We mainly apply SeDA to post-attention linear layers (e.g., MLP), as they directly precede residual addition and heavily influence layer outputs. After quantizing the target layer, we perform bin-count profiling on anchor tokens to measure how often each dialect is selected in each dynamic range. We then construct an 8-dialect sub-formatbook by selecting the most frequent dialect for each dynamic-range bucket (Figure~\ref{fig:seda}b), and apply it to constrain the semantically related tokens at the same timestep.

\para{Attention-Guided Semantic Relatedness Extraction}
Prior works use attention scores as a criterion for token similarity~\citep{lu2025toma} and show that VDiT attention map exhibits strong spatiotemporal locality and encodes prompt-conditioned structure~\citep{cai2025ditctrl, wen2025analysis}. Following this line, we treat attention scores ($Q\cdot K^T$) as a \emph{proxy} for semantic relatedness.
Concretely, we identify \emph{anchor tokens} and their \emph{correlated tokens}---tokens that receive strong attention from the anchor---and constrain them to share the same sub-formatbook (Figure~\ref{fig:seda}a).

Since selecting anchors by global top-$k$ attention score is computationally expensive and suffers from poor spatial coverage (clustering in large homogeneous regions such as background), we propose a localized selection strategy.
For \emph{factorized attention}, \circnum{1} we partition each frame into spatial tiles and identify one anchor candidate per frame per tile ($N$ candidates per tile, $N$: \# frames) with the highest mean attention within each tile; \circnum{2} we select a global main anchor from these $N$ candidates using the highest temporal mean attention score and prune frames whose candidates are weakly correlated with the main anchor; and \circnum{3} within a local square window centered at the anchor candidates of the unpruned frames, we mark tokens whose attention scores from the anchor exceed a threshold as correlated tokens. For \emph{3D attention}, we directly select the main anchor as the token with the highest mean attention score over $k \times k \times N$ tokens ($k \times k$ spatial tile across $N$ frames), then mark tokens within a local cuboid window whose attention score from the anchor exceeds a threshold as correlated tokens. 
In both cases, raw attention scores are used to exclude weakly or negatively related tokens.

\para{Mitigating Profiling Overhead}
Although SeDA is applied to only a small number of layers, bin-count profiling and anchor/correlated token identification can still be expensive. Figure~\ref{fig:seda}c reveals a key insight: anchor token sets exhibit a U-shaped stability pattern across denoising timesteps. In early timesteps, the attention map is unstable, leading to high variance; we therefore skip SeDA in this region. In late timesteps, where video details are refined, updating tokens at every timestep is beneficial. For the intermediate stable region, we perform periodic updates (e.g., every 10 timesteps), which cause little quality degradation. These strategies collectively minimize SeDA's runtime overhead throughout the denoising process. Ablation results are in Appendix~\ref{app:seda_ablation}.


\section{Experiments}
\label{sec:experiments}

\begin{table}[t]
\caption{Text-to-video generation quality on VBench benchmark suite for Open-Sora~1.0 and 2.0.
}\label{tab:main}
\renewcommand{\arraystretch}{1.0}

\centering
\scriptsize
\setlength{\tabcolsep}{3pt}
\begin{tabular}{c|ccc|cccccccc}
\toprule

\shortstack{\textbf{Model}\\\ } & \shortstack{\textbf{Method}\\\ } & \shortstack{\textbf{Block}\\[-1pt]\ \textbf{Size}} & \shortstack{\textbf{Eff. bit}\\[-1pt]\textbf{(A/W)}} & \shortstack{\textbf{Aesthetic}\\[-1pt]\textbf{Quality}} & \shortstack{\textbf{Imaging}\\[-2pt]\textbf{Quality}} & \shortstack{\textbf{Motion}\\[-1pt]\textbf{Smooth.}} & \shortstack{\textbf{Dynamic}\\[-2pt]\textbf{Degree}} & \shortstack{\textbf{Subject}\\[-2pt]\textbf{Consist.}} & \shortstack{\textbf{BG.}\\[-1pt]\textbf{Consist.}} & \shortstack{\textbf{Scene}\\[-1pt]\textbf{Consist.}} & \shortstack{\textbf{SS.}\\[-1pt]\textbf{ Consist.}} \\
\hline  \rule{0pt}{2.0ex}

\multirow{13}{*}{\shortstack{\textbf{Open-Sora}\\\ \textbf{1.0}\\\ factorized \\\ attn}}
& FP16 & - & 16 / 16 & 58.00 & 62.03 & 96.41 & 54.17 & 91.51 & 96.43 & 36.17 & 26.14\\
\cline{2-12} \rule{0pt}{2.0ex}
& \multirow{2}{*}{MXFP4} & 16 & 4.31 / 4.31 & 33.22 & 30.78 & 98.74 & 82.87 & 85.24 & 94.28 & 4.70 & 14.39 \\
&  & 32 & 4.16 / 4.16 & 34.69 & 37.80 & 98.59 & 84.72 & 85.99 & 94.53 & 5.69 & 15.00 \\
\cline{2-12}  \rule{0pt}{2.0ex}
& \multirow{2}{*}{NVFP4} & 16 & 4.5 / 4.5 & 51.41 & 52.46 & 97.29 & 56.94 & 88.77 & 95.50 & 16.28 & 23.17 \\
&  & 32 & 4.25 / 4.25 & 46.82 & 47.26 & 97.90 & 50.00 & 88.80 & 95.61 & 13.42 & 20.88 \\
\cline{2-12}  \rule{0pt}{2.0ex}
& \multirow{2}{*}{\shortstack{Block\\Dialect}} & 16 & 4.56 / 4.56 & 49.03 & 47.65 & 98.02 & 58.80 & 88.25 & 95.12 & 15.72 & 21.97 \\
&  & 32 & 4.28 / 4.28 & 44.27 & 46.14 & 98.33 & 68.06 & 87.71 & 94.92 & 9.08 & 19.72 \\
\cline{2-12}  \rule{0pt}{2.0ex}

& ViDiT-Q & \multirow[c]{2}{*}{\shortstack[c]{\strut A: Token\\[-3pt]W: Ch\strut}} & \multirow{2}{*}{\shortstack[c]{4.81 / 4.73\\[-3pt]\ }} & 30.71 & 62.73 & 82.43 & 45.37 & 91.61 & 95.08 & 0.39 & 8.33 \\
& Q-VDiT &  &  & 28.72 & 53.98 & 84.91 & 86.57 & 97.01 & 97.37 & 0.00 & 2.08 \\

\hhline{~-----------}  \rule{0pt}{2.0ex}
& \cellcolor{gray!15} & \cellcolor{gray!15}16 & \cellcolor{gray!15}4.63 / 4.63 & \cellcolor{gray!15}51.83 & \cellcolor{gray!15}54.84 & \cellcolor{gray!15}97.63 & \cellcolor{gray!15}42.59 & \cellcolor{gray!15}90.99 & \cellcolor{gray!15}96.21 & \cellcolor{gray!15}25.73 & \cellcolor{gray!15}23.80 \\
& \cellcolor{gray!15}\multirow{-2}{*}{\raisebox{-1pt}{\formatname}} & \cellcolor{gray!15}32 & \cellcolor{gray!15}4.31 / 4.31 & \cellcolor{gray!15}45.95 & \cellcolor{gray!15}48.69& \cellcolor{gray!15}98.31 & \cellcolor{gray!15}47.22 & \cellcolor{gray!15}88.90 & \cellcolor{gray!15}95.60 & \cellcolor{gray!15}10.63 & \cellcolor{gray!15}20.98 \\
\cline{2-12}  \rule{0pt}{2.0ex}
& \cellcolor{gray!25} \raisebox{-1pt}{Semantic}& \cellcolor{gray!25}16 & \cellcolor{gray!25}5.10 / 4.63 & \cellcolor{gray!25}54.00 & \cellcolor{gray!25}59.52 & \cellcolor{gray!25}97.09 & \cellcolor{gray!25}44.44 & \cellcolor{gray!25}91.57 & \cellcolor{gray!25}95.88 & \cellcolor{gray!25}32.15 & \cellcolor{gray!25}24.64 \\
& \cellcolor{gray!25}Dialect & \cellcolor{gray!25}32 & \cellcolor{gray!25}4.76 / 4.31 & \cellcolor{gray!25}51.09 & \cellcolor{gray!25}57.04 & \cellcolor{gray!25}97.40 & \cellcolor{gray!25}51.39 & \cellcolor{gray!25}89.48 & \cellcolor{gray!25}95.22 & \cellcolor{gray!25}20.93 & \cellcolor{gray!25}22.94 \\
\hline\hline  \rule{0pt}{2.0ex}

\multirow{13}{*}{\shortstack{\textbf{Open-Sora}\\\ \textbf{2.0}\\\ 3D attn}}
& FP16 & - & 16 / 16 & 57.35 & 63.26 & 98.86 & 62.50 & 94.79 & 97.98 & 48.76 & 27.98 \\
\cline{2-12}  \rule{0pt}{2.0ex}
& \multirow{2}{*}{MXFP4} & 16 & 4.31 / 4.31 & 50.31 & 39.40 & 98.98 & 53.24 & 86.87 & 94.98 & 36.60 & 26.00 \\
&  & 32 & 4.16 / 4.16 & 51.11 & 41.84 & 98.99 & 54.17 & 87.26 & 95.15 & 39.97 & 26.25 \\
\cline{2-12}  \rule{0pt}{2.0ex}
& \multirow{2}{*}{NVFP4} & 16 & 4.5 / 4.5 & 56.24 & 62.44 & 98.60 & 62.96 & 92.21 & 97.00 & 48.74 & 28.07 \\
&  & 32 & 4.25 / 4.25 & 55.74 & 60.33 & 98.55 & 65.28 & 91.22 & 96.43 & 46.49 & 27.97 \\
\cline{2-12}  \rule{0pt}{2.0ex}
& \multirow{2}{*}{\shortstack{Block\\Dialect}} & 16 & 4.56 / 4.56 & 56.83 & 62.22 & 98.72 & 59.72 & 93.09 & 97.40 & 48.04 & 28.12 \\
&  & 32 & 4.28 / 4.28 & 56.41 & 61.00 & 98.70 & 61.57 & 92.53 & 96.89 & 47.55 & 28.08 \\
\cline{2-12}  \rule{0pt}{2.0ex}
& \multirow[c]{2}{*}{\shortstack[c]{ViDiT-Q\\[-4.5pt]}}
& \multirow[c]{2}{*}{\shortstack[c]{\strut A: Token\\[-3pt]W: Ch\strut}}
& \multirow[c]{2}{*}{\raisebox{-0.3ex}{4.82 / 4.40}} & \multirow[c]{2}{*}{54.15} & \multirow[c]{2}{*}{58.01} & \multirow[c]{2}{*}{98.74} & \multirow[c]{2}{*}{52.31} & \multirow[c]{2}{*}{89.43} & \multirow[c]{2}{*}{94.81} & \multirow[c]{2}{*}{41.33} & \multirow[c]{2}{*}{27.30} \\
&  &  &  &  &  &  &  &  &  &  \\

\hhline{~-----------}  \rule{0pt}{2.0ex}
& \cellcolor{gray!15} & \cellcolor{gray!15}16 & \cellcolor{gray!15}4.63 / 4.63 & \cellcolor{gray!15}56.82 & \cellcolor{gray!15}63.12 & \cellcolor{gray!15}98.66 & \cellcolor{gray!15}61.57 & \cellcolor{gray!15}93.24 & \cellcolor{gray!15}97.29 & \cellcolor{gray!15}48.35 & \cellcolor{gray!15}28.13 \\
& \cellcolor{gray!15}\multirow{-2}{*}{\raisebox{-1pt}{\formatname}} & \cellcolor{gray!15}32 & \cellcolor{gray!15}4.31 / 4.31 & \cellcolor{gray!15}56.48 & \cellcolor{gray!15}61.66 & \cellcolor{gray!15}98.66 & \cellcolor{gray!15}64.81 & \cellcolor{gray!15}92.62 & \cellcolor{gray!15}96.86 & \cellcolor{gray!15}48.21 & \cellcolor{gray!15}28.07 \\
\cline{2-12}  \rule{0pt}{2.0ex}

& \cellcolor{gray!25} \raisebox{-1pt}{Semantic}& \cellcolor{gray!25}16 & \cellcolor{gray!25}4.63 / 4.63 & \cellcolor{gray!25}56.50 & \cellcolor{gray!25}63.75 & \cellcolor{gray!25}98.59 & \cellcolor{gray!25}67.59 & \cellcolor{gray!25}93.02 & \cellcolor{gray!25}96.90 & \cellcolor{gray!25}49.54 & \cellcolor{gray!25}28.14 \\
& \cellcolor{gray!25}Dialect & \cellcolor{gray!25}32 & \cellcolor{gray!25}4.31 / 4.31 & \cellcolor{gray!25}56.55 & \cellcolor{gray!25}63.24 & \cellcolor{gray!25}98.51 & \cellcolor{gray!25}66.20 & \cellcolor{gray!25}92.34 & \cellcolor{gray!25}96.31 & \cellcolor{gray!25}48.57 & \cellcolor{gray!25}28.27 \\
\bottomrule
\end{tabular}
\end{table}

\subsection{Experimental Setup}
\para{Models and Evaluation Settings}
We evaluate \papername on Open-Sora~1.0 (724M)~\citep{zheng2024open} and Open-Sora~2.0 (11B)~\citep{peng2025open2}, covering two representative architectures: factorized spatial/temporal attention and full 3D attention, respectively. For Open-Sora~1.0, we use 100 denoising steps with CFG scale 4.0; Open-Sora~2.0 uses 50 steps with CFG scale 7.5 (text) and 3.0 (image). We employ VBench~\citep{huang2024vbench} and report eight key dimensions, following prior works~\citep{ren2024consisti2v,feng2025q}. We abbreviate \textit{Background Consistency} as \textit{BG Consist.} and \textit{Semantic \& Style Consistency} as \textit{SS. Consist.} Ablation studies use a reduced prompt set for computational efficiency. In addition, following EvalCrafter~\citep{liu2024evalcrafter}, we report CLIPSIM for text--video alignment and CLIP-Temp for temporal consistency, DOVER~\citep{wu2023exploring} for aesthetic and technical quality (VQA-A/T), and flow score for motion information. Details are in Appendices~\ref{app:metrics} and \ref{app:eval_setting}. 

\para{Baseline}
We compare \papername against representative block-wise formats: MXFP4~\citep{ocp_mx_v1}, which uses power-of-two block scaling with a custom 5-bit exponent (covering the FP16 range), and NVFP4~\citep{nvidia_nvfp4_blog}, which uses FP8 block scaling with additional FP32 per-tensor scales. We also compare against two VDiT quantization schemes: ViDiT-Q~\citep{zhao2024vidit}, which combines fine-grained grouping, channel balancing, and mixed precision; and Q-VDiT~\citep{feng2025q}, which applies low-rank approximation of quantization error and inter-frame distribution distillation. Q-VDiT is compared only on Open-Sora~1.0 as its implementation requires learning parameters for isolated temporal attention. We additionally compare against BlockDialect~\citep{jangblockdialect}, a mixed-format quantization that also performs per-block dialect selection but uses a 16-dialect formatbook and a different dialect-selection strategy. \papername uses \emph{num\_groups}=8 by default. For all block-wise methods, we use block sizes of 16 and 32. For ViDiT-Q and Q-VDiT, we apply their mixed-precision settings for comparable effective bitwidths.


\subsection{Main Results}
Table~\ref{tab:main} summarizes the main results. On Open-Sora~1.0, prior VDiT quantization methods and MXFP4 often fail to generate recognizable videos in the sub-5-bit activation regime, as evidenced by abnormally elevated dynamic degree scores caused by noise and degraded scene consistency. While BlockDialect and NVFP4 produce better results, \papername achieves superior aesthetic and imaging quality with enhanced scene consistency. On Open-Sora~2.0, \papername still broadly outperforms or matches all baselines, approaching FP16 quality. Note that \papername operates at a higher effective bitwidth on Open-Sora~1.0 than 2.0 due to matrix activation decomposition. Notably, \formatname alone (without decomposition and SeDA) already shows strong performance against NVFP4 across most metrics; additional metric comparisons are provided in Appendix~\ref{app:nvfp4}. \papername also proves compatible with other PTQ methods, as demonstrated by additional gains when combined with rotation-based PTQ (Appendix~\ref{app:rotation}).

\begin{table}[t]
\centering
\caption{Dialect selection strategy comparison (LUT-based vs.\ MSE-based). LUT-based selection matches or outperforms MSE-based strategy. We denote \emph{num\_group} by $g$. }
\label{tab:exact_mse_vs_group_selection}
\setlength{\tabcolsep}{3pt}
\renewcommand{\arraystretch}{0.9}
\fontsize{7.7}{8.8}\selectfont   
\begin{tabular}{llccccccccc}
\toprule 

\multicolumn{1}{c}{\shortstack[c]{\textbf{Model}\\[1pt]\ }} & \multicolumn{1}{c}{\shortstack[c]{\textbf{Method}\\[1pt]\ }} & \shortstack{\textbf{Eff. bit}\\[-1pt]\textbf{(A/W)}} & \shortstack{\textbf{Aesthetic}\\[-1pt]\textbf{Quality}} & \shortstack{\textbf{Imaging}\\[-2pt]\textbf{Quality}} & \shortstack{\textbf{Motion}\\[-1pt]\textbf{Smooth.}} & \shortstack{\textbf{Dynamic}\\[-2pt]\textbf{Degree}} & \shortstack{\textbf{Subject}\\[-2pt]\textbf{Consist.}} & \shortstack{\textbf{BG.}\\[-1pt]\textbf{Consist.}} & \shortstack{\textbf{Scene}\\[-1pt]\textbf{Consist.}} & \shortstack{\textbf{SS.}\\[-1pt]\textbf{ Consist.}} \\
\hline  
\rule{0pt}{2.0ex} \multirow{3}{*}{\shortstack[c]{\textbf{Open-Sora}\\\textbf{1.0}}}
\rule{0pt}{2.0ex}
& Exact MSE       &           & 45.84 & 50.44 & 98.40 & 45.83 & 90.71 & 95.78 & 9.52 & 19.94 \\ 
& \cellcolor{gray!10}\formatname (g=32)  & 4.31/4.31 & \cellcolor{gray!10}45.97 & \cellcolor{gray!10}50.61 & \cellcolor{gray!10}98.37 & \cellcolor{gray!10}48.61 & \cellcolor{gray!10}90.46 & \cellcolor{gray!10}95.63 & \cellcolor{gray!10}10.12 & \cellcolor{gray!10}19.77 \\
& \cellcolor{gray!20}\formatname (g=8)  &           & \cellcolor{gray!20}45.06 & \cellcolor{gray!20}50.15 & \cellcolor{gray!20}98.41 & \cellcolor{gray!20}44.44 & \cellcolor{gray!20}90.68 & \cellcolor{gray!20}95.41 & \cellcolor{gray!20}9.60 & \cellcolor{gray!20}19.73 \\
\hline \rule{0pt}{2.0ex}
\multirow{3}{*}{\shortstack[c]{\textbf{Open-Sora}\\\textbf{2.0}}}
\rule{0pt}{2.0ex}
& Exact MSE       &           & 57.52 & 64.01 & 98.67 & 59.72 & 92.32 & 97.01 & 40.63 & 27.46 \\
& \cellcolor{gray!10}\formatname (g=32)  & 4.31/4.31 & \cellcolor{gray!10}57.62 & \cellcolor{gray!10}63.31 & \cellcolor{gray!10}98.67 & \cellcolor{gray!10}54.17 & \cellcolor{gray!10}92.35 & \cellcolor{gray!10}97.03 & \cellcolor{gray!10}42.63 & \cellcolor{gray!10}27.54 \\
& \cellcolor{gray!20}\formatname (g=8)  &           & \cellcolor{gray!20}57.44 & \cellcolor{gray!20}63.80 & \cellcolor{gray!20}98.69 & \cellcolor{gray!20}58.33 & \cellcolor{gray!20}92.07 & \cellcolor{gray!20}96.81 & \cellcolor{gray!20}40.63 & \cellcolor{gray!20}27.54 \\
\bottomrule
\end{tabular}
\end{table}

\subsection{Comparison with MSE-based Dialect Selection}
Table~\ref{tab:exact_mse_vs_group_selection} compares LUT-based and MSE-based dialect selection strategies at block size 32, with two observations. First, \formatname (\emph{num\_groups}=32) with LUT-based dialect selection outperforms exact MSE-based selection, suggesting that MSE-driven optimization, often dominated by a few extreme outliers, does not necessarily translate to superior end-to-end video quality. Second, even coarse group-wise maximum extraction (\emph{num\_groups}=8) remains competitive with exact MSE selection, indicating that a rough approximation of large-magnitude values is sufficient to guide effective dialect selection. \papername uses a default \emph{num\_groups} of 8 to balance performance and efficiency.



\begin{table}[t]
\centering
\caption{Stepwise ablation on Open-Sora 1.0 (OpenSora t2v\_sample prompt set).}
\label{tab:progressive_improvement_seda}
\setlength{\tabcolsep}{3pt}
\renewcommand{\arraystretch}{0.8}
\fontsize{7.7}{8.8}\selectfont   
\begin{tabular}{lccccccc}
\toprule
\shortstack[c]{\textbf{Method} } &
\shortstack[c]{\textbf{Block size}} &
\shortstack[c]{\textbf{FVD-FP16} ($\downarrow$) } &
\shortstack[c]{\textbf{CLIP-Temp} } &
\shortstack[c]{\textbf{CLIPSIM}} &
\shortstack[c]{\textbf{VQA-A}} &
\shortstack[c]{\textbf{VQA-T}} &
\shortstack[c]{\textbf{$\Delta$ Flow}($\downarrow$)} \\
\midrule
FP16  & -- & --    & 0.9987 & 0.1798 & 51.66 & 48.39 & --  \\
\midrule

\formatname   &   \multirow{3}{*}{16}    &  1.24 & 0.9983 & 0.1781 & 50.72 & 47.87 & 0.26 \\
+Decomp. & &  0.80 & 0.9989 & 0.1751 & 51.01 & 48.14 & 0.31 \\
+SeDA &     &  0.76 & 0.9988 & 0.1757 & 51.03 & 48.10 & 0.26 \\
\midrule

\formatname   &  \multirow{3}{*}{32}   &  1.59 & 0.9986 & 0.1778 & 50.29 & 47.78 & 0.84 \\
+Decomp. & &  1.36 & 0.9986 & 0.1758 & 50.88 & 48.06 & 0.55 \\
+SeDA &     &  1.26 & 0.9983 & 0.1763 & 50.86 & 48.05 & 0.44 \\
\bottomrule
\end{tabular}
\end{table}

\subsection{Stepwise Evaluation with Additional Metrics}
To analyze the contribution of each proposed technique, we conduct a stepwise ablation on Open-Sora~1.0 using the OpenSora \texttt{t2v\_sample} prompt set, averaged over multiple random seeds. As shown in Table~\ref{tab:progressive_improvement_seda}, FVD-FP16, which measures feature-distribution similarity to the FP16 baseline and correlates highly with human perception, consistently improves as more techniques are incorporated. Similarly, $\Delta$Flow score, which captures the gap in motion intensity relative to the FP16 baseline, is minimized when all techniques are applied. Across most metrics, the most substantial gain occurs upon applying activation decomposition, highlighting the importance of handling quantization-sensitive layers. Additional results on different datasets are provided in Appendix~\ref{app:cumulative}.

\begin{figure}[t]
  \centering
  \includegraphics[page=15, width=0.9\columnwidth]{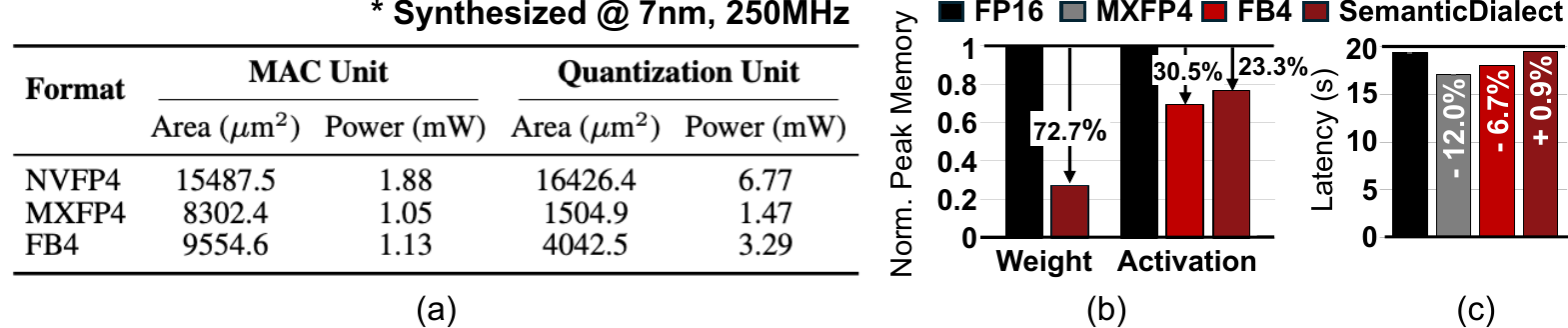}

  
  \caption{Hardware deployment analysis. (a) RTL: MAC and quantization unit comparison. (b) CUDA: Peak memory. (c) CUDA: DiT block inference latency (projected Blackwell-class latency).}
  \label{fig:hardware}
\end{figure}
\subsection{Hardware Deployment Analysis}
\label{sec:hardware}
\para{Implementation} 
We implement \papername in RTL and CUDA to demonstrate its deployment practicality. For RTL, we extend FlexASR~\citep{tambe202216} on TSMC 7nm at 250MHz (Appendix~\ref{app:hardware}.1). For GPU kernel, \formatname's normalized integer values (0--15) are incompatible with native MXFP4 kernels but representable in FP8 e4m3, making them MXFP8-compatible. As H100 is our only available GPU and does not natively support MXFP8 GEMM, we estimate Blackwell-class latency by halving FP16 GEMM latency, reflecting MXFP8's 2$\times$ throughput (Appendix~\ref{app:hardware}.2). We also implement an MXFP4 kernel (same power-of-two scaling with FP4) assuming the same MXFP8 GEMM to isolate \formatname quantization overhead, not to compare latency against native 4-bit hardware.

\para{RTL Results}
\formatname achieves notably higher hardware efficiency than NVFP4 (Figure~\ref{fig:hardware}a). NVFP4's inefficiency stems from FP8 block-wise and FP32 tensor-wise scaling, which require extra multipliers and complicate FSM logic, whereas \formatname and MXFP4 reduce scaling to simple exponent arithmetic. While MXFP4 shows marginally better efficiency, it substantially degrades generation quality. Formatbook area of FB4 MAC accounts for only 12.6\% of multipliers (excluding accumulators).

\para{CUDA Kernel Results}
\emph{Peak Memory:} \papername reduces weight and activation peak memory by 72.7\% and 23.3\%, respectively (Figure~\ref{fig:hardware}b). The moderate activation reduction is due to the co-existence of quantized and non-quantized tensors (e.g., residual connections, GeLU inputs). Without SeDA, activation memory reduction improves to 30.5\%. \emph{Inference Latency:} Figure~\ref{fig:hardware}c shows that MXFP4 (via MXFP8 GEMM) achieves 12.0\% latency reduction over FP16, reasonable given $\sim$17.5\% MXFP8 gains on Blackwell~\citep{pytorch_blog}. \formatname's online quantization introduces only 5.9\% latency overhead over MXFP4, and the full \papername pipeline incurs only 0.9\% overhead over FP16, with room for further reduction via kernel fusion and native Blackwell support (Appendix~\ref{app:hardware}.2).

\section{Conclusion}
\label{sec:conclusion}
In this work, we propose \papername, a PTQ method for VDiTs that improves generation quality under 4-bit quantization via LUT-based dialect selection, attention-guided activation decomposition, and SeDA for spatiotemporal consistency. Experiments demonstrate that \papername consistently outperforms prior VDiT quantization methods and block-wise format baselines, showing that accurate edge deployment is achievable with scalable mixed-format quantization. 
\textbf{Limitation}: While this work demonstrates the effectiveness of semantic-aware mixed-format quantization, the configuration space of SeDA---including anchor and correlated token selection, sub-formatbook design, and timestep scheduling---may not be globally optimized. Future work could explore tailoring these configurations to video-specific characteristics such as motion intensity.

\newpage
\bibliographystyle{plainnat} 
\bibliography{references}

\newpage
\appendix
\label{sec:appendix}

\section{Layer-wise Quantization Sensitivity Test}
\label{app:sensitivity}

\begin{table}[H]
\caption{Open-Sora 1.0: per-layer sensitivity. Top: FP16 baseline with only one layer quantized. Bottom: \formatname baseline with only one layer set to FP16. \textbf{Bold}: notable degradation or improvement.}
\label{tab:opensora1_layer_sensitivity}
\renewcommand{\arraystretch}{0.9} 

\centering
\small
\setlength{\tabcolsep}{4.5pt}
\begin{tabular}{lcccccccc}
\toprule
  & \shortstack{Aesthetic\\quality}
 & \shortstack{Imaging\\quality}
 & \shortstack{Motion\\smooth.}
 & \shortstack{Dynamic\\degree}
 & \shortstack{Subject\\consist.}
 & \shortstack{BG.\\consist.}
 & \shortstack{Scene\\consist. }
 & \shortstack{SS.\\consist.} \\
\midrule
FP16 &  58.53 & 65.33 & 97.29 & 47.22 & 93.04 & 95.98 & 25.00 & 25.28\\
attn\_temp.qkv & \textbf{56.64} & \textbf{59.63} & 97.90 & 44.44 & 93.34 & 96.26 & 25.00 & 25.05 \\
attn\_temp.proj & 58.47 & 65.55 & 97.27 & 47.22 & 93.12 & 95.88 & 27.90 & 25.21 \\
attn.qkv & 57.32 & 64.44 & 97.21 & 58.33 & 92.29 & 95.37 & 29.09 & 25.27 \\
attn.proj & 58.37&	65.15&	97.18&	45.83&	93.07&	95.95&	25.30&	25.62 \\
x\_attn.q\_linear & 58.92&	64.99&	97.04&	47.22&	92.44	&95.93&	28.94&	25.16 \\
x\_attn.kv\_linear & 58.99&	66.11&	97.23&	51.39&	93.11&	96.10&	28.05&	25.25 \\
x\_attn.proj & 59.47&	65.88&	97.27&	48.61&	93.07&	96.19&	27.08&	25.46 \\
mlp.fc1 & 57.55&	65.48&	97.13&	50.00&	\textbf{91.80}&	95.87&	26.71&	24.67 \\
mlp.fc2 & \textbf{54.79}&	\textbf{60.12}&	98.59&	\textbf{29.17}&	93.69&	96.60&	\textbf{18.38}&	\textbf{24.19} \\
\midrule
\formatname & 45.06 & 50.15 & 98.41 & 44.44 & 90.68 & 95.41 & 9.60 & 19.73 \\
attn\_temp.qkv & \textbf{49.23} & \textbf{58.51} & 98.02 & 47.22 & 91.06 & 95.51 & \textbf{14.14} & \textbf{21.62} \\
attn\_temp.proj & 46.12 & 50.85 & 98.45 & 37.50 & 90.98 & 95.60 & 9.97 & 20.21 \\
attn.qkv & 47.30 & 51.62 & 98.43 & 36.11 & 90.96 & 95.91 & 13.62 & 20.96 \\
attn.proj & 46.38 & 49.90 & 98.48 & 40.28 & 90.66 & 95.78 & 11.76 & 20.69 \\  
x\_attn.q\_linear & 45.65 & 51.65 & 98.50 & 43.06 & 91.06 & 95.67 & 11.38 & 20.15 \\
x\_attn.kv\_linear & 45.44 & 50.76 & 98.36 & 40.28 & 90.51 & 95.44 & 11.16 & 20.56 \\
x\_attn.proj & 45.01 & 50.00 & 98.50 & 47.22 & 90.72 & 95.30 & 11.68 & 19.72 \\
mlp.fc1 & 46.44 & 51.25 & 98.45 & 38.89 & 90.99 & 96.09 & 13.47 & 20.85 \\
mlp.fc2 & \textbf{51.05} & \textbf{53.37} & 97.61 & 44.44 & 90.23 & 95.28 & \textbf{20.68} & \textbf{22.32} \\
\bottomrule
\end{tabular}

\end{table}


\begin{table}[H]
\caption{Open-Sora 2.0: per-layer sensitivity. Top: FP16 baseline with only one layer quantized. Bottom: \formatname baseline with only one layer set to FP16. \textbf{Bold}: notable degradation or improvement.}
\label{tab:opensora2_layer_sensitivity}
\renewcommand{\arraystretch}{0.9} 

\centering
\small
\setlength{\tabcolsep}{4.5pt}
\begin{tabular}{lcccccccc}
\toprule
 & \shortstack{Aesthetic\\quality}
 & \shortstack{Imaging\\quality}
 & \shortstack{Motion\\smooth.}
 & \shortstack{Dynamic\\degree}
 & \shortstack{Subject\\consist.}
 & \shortstack{BG.\\consist.}
 & \shortstack{Scene\\consist. }
 & \shortstack{SS.\\consist.} \\
\midrule
FP16 & 58.05 & 64.58 & 98.83 & 59.72 & 94.43 & 97.80 & 40.85 & 27.37 \\

img\_attn.qkv & 57.74 & 64.51 & 98.80 & 56.94 & 93.83 & 97.74 & 41.07 & 27.40 \\
img\_attn.proj & 57.91 & 64.64 & 98.83 & 55.56 & 94.21 & 97.83 & 41.74 & 27.46 \\
img\_mlp.0 & \textbf{57.35} & 64.81 & 98.63 & 59.72 & 93.51 & 97.27 & 41.44 & 27.61 \\
img\_mlp.2 & \textbf{57.41} & 64.85 & 98.79 & 55.56 & 94.27 & 97.76 & 41.89 & 27.49 \\
img\_mod.lin & 57.90 & 65.31 & 98.71 & 54.17 & 93.65 & 97.57 & 40.63 & 27.42 \\
txt\_attn.qkv & 58.07 & 64.62 & 98.82 & 61.11 & 94.39 & 97.87 & 40.03 & 27.36 \\
txt\_attn.proj & 58.07 & 64.76 & 98.83 & 56.94 & 94.23 & 97.62 & 40.48 & 27.38 \\
txt\_mlp.0 & 57.85 & 64.67 & 98.84 & 58.33 & 94.24 & 97.76 & 41.15 & 27.43 \\
txt\_mlp.2 & 58.01 & 64.51 & 98.79 & 55.56 & 94.03 & 97.81 & 43.30 & 27.34 \\
txt\_mod.lin & 58.08 & 64.46 & 98.80 & 56.94 & 94.10 & 97.62 & 41.67 & \textbf{27.19} \\
linear1  & 58.09 & 64.46 & 98.78 & \textbf{51.39} & 93.95 & 97.93 & \textbf{39.06} & 27.33 \\
linear2 & 58.19 & 64.86 & 98.74 & 55.56 & 94.07 & 97.73 & 40.85 & 27.51 \\
modulation.lin & 58.49 & \textbf{61.84} & 98.86 & 61.11 & 94.06 & 97.92 & 39.58 & 27.33 \\
\midrule

\formatname & 57.44 & 63.80 & 98.69 & 58.33 & 92.07 & 96.81 & 40.63 & 27.54 \\

img\_attn.qkv & 57.74 & 63.35 & 98.64 & 54.17 & 92.19 & 96.97 & 42.04 & 27.63 \\
img\_attn.proj & 57.60 & 63.46 & 98.66 & 56.94 & 92.38 & 96.77 & 40.55 & 27.62 \\
img\_mlp.0 & 57.63 & 63.28 & 98.73 & 59.72 & 92.87 & 97.28 & 42.78 & 27.51 \\
img\_mlp.2 & 57.50 & 63.58 & 98.69 & 55.56 & 92.29 & 97.05 & 41.89 & 27.41 \\
img\_mod.lin & 57.66 & 62.49 & 98.68 & 52.78 & 92.12 & 97.16 & 41.59 & 27.64 \\
txt\_attn.qkv & 57.60 & 63.99 & 98.63 & 58.33 & 92.22 & 96.65 & 42.56 & 27.70 \\
txt\_attn.proj & 57.56 & 63.79 & 98.67 & 55.56 & 92.27 & 96.98 & 41.89 & 27.59 \\
txt\_mlp.0 & 57.42 & 63.75 & 98.66 & 58.33 & 92.19 & 96.84 & 43.53 & 27.63 \\
txt\_mlp.2 & 57.40 & 63.92 & 98.68 & 63.89 & 92.02 & 96.80 & 43.15 & 27.70 \\
txt\_mod.lin & 57.33 & 63.49 & 98.56 & 55.56 & 92.16 & 96.68 & 37.35 & 27.57 \\
linear1  & 57.69 & 64.13 & 98.65 & 62.50 & 92.30 & 96.86 & 39.51 & 27.49 \\
linear2 & 57.28 & 63.49 & 98.78 & 55.56 & 92.35 & 96.95 & 41.59 & 27.48 \\
modulation.lin & 57.64 & \textbf{65.42} & 98.53 & 54.17 & 91.84 & 96.09 & \textbf{44.05} & \textbf{27.73} \\
\bottomrule
\end{tabular}

\end{table}

Tables~\ref{tab:opensora1_layer_sensitivity} and~\ref{tab:opensora2_layer_sensitivity} summarize the layer-wise quantization sensitivity of Open-Sora~1.0 and~2.0. In Open-Sora~1.0, \texttt{attn\_temp.qkv} and \texttt{mlp.fc2} are clearly the most sensitive layers. In the larger Open-Sora~2.0, the model is generally more robust to quantization and exhibits smaller layer-wise variation, but \texttt{modulation.lin} (the modulation layer in \texttt{SingleProcessBlock}) remains the most sensitive. 
Additonally, we prioritize \texttt{txt\_mod.lin} (the text modulation layer in \texttt{DoubleProcessBlock}) for activation decomposition because its vector-shaped activation makes decomposition computationally inexpensive, and it empirically shows strong results when decomposed.


\begin{figure}[t]
  \centering
  \includegraphics[page=5, width=0.9\columnwidth]{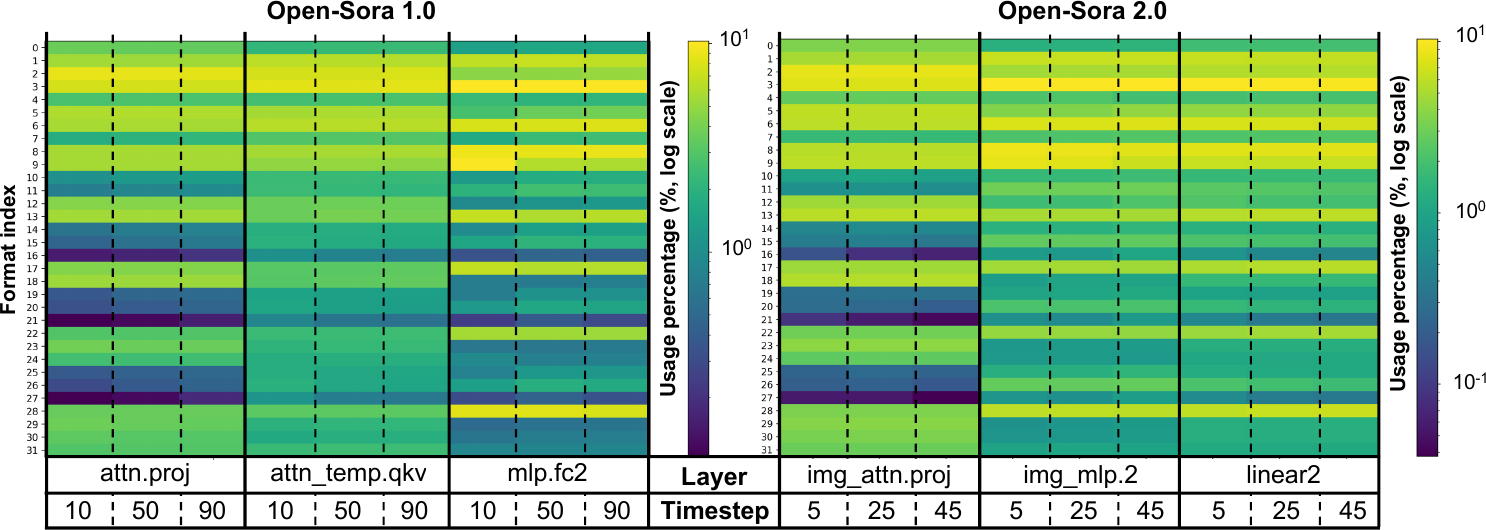}
\caption{Block-wise format usage.}
  \label{fig:format_usage}
\end{figure}
\section{Block-wise Format Usage}
\label{app:format_usage}
Figure~\ref{fig:format_usage} reports the average format usage across layers and denoising timesteps, averaged over 10 prompts. We observe clear layer-specific preferences, but no format collapse (i.e., every format is selected at least occasionally). The least-used formats (Formats 16, 21, and 27) appear at around 0.1\% in the projection layer immediately after spatial attention, yet are meaningfully used in other layers. This indicates that the full format set is leveraged when beneficial. Overall, lower-index formats, which provide a narrower dynamic range (smaller block maxima), are selected more often, consistent with the predominance of small-magnitude normalized values. Importantly, usage patterns depend primarily on layer type rather than denoising timestep. For example, in Open-Sora 1.0, temporal attention QKV projection (\texttt{attn\_temp.qkv}) shows relatively uniform usage, whereas the final MLP layer (\texttt{mlp.fc2}) consistently concentrates on a few formats (e.g., Formats 28, 22, and 17) even within the same dynamic range group. These results suggest that block-wise mixed formats systematically adapt to each layer without explicit calibration.

\section{Evaluation Metrics Description}
\label{app:metrics}

\subsection{VBench Benchmark.}
Following VBench~\citep{huang2024vbench} and prior VDiT quantization work~\citep{zhao2024vidit, feng2025q}, we evaluate on 8 VBench dimensions spanning three aspects of video generation: \textbf{(i) Frame-wise Quality}, \textbf{(ii) Temporal Quality}, and \textbf{(iii) Semantics}.
We use the official VBench prompt sets below and generate three videos per prompt using different random seeds to improve stability. For the main results, we use the full prompt set, while for other evaluations we subsample one-third of the prompts within each category to reduce inference cost.

\begin{itemize}
    \item \texttt{overall\_consistency.txt} (93 prompts) for semantic \& style consistency, aesthetic quality and imaging quality
    \item \texttt{subject\_consistency.txt} (72 prompts) for subject consistency, dynamic degree and motion smoothness
    \item \texttt{scene.txt} (86 prompts) for background consistency and scene consistency
\end{itemize}

\subsection{VBench dimensions}
\textbf{Frame-wise Quality} measures per-frame visual quality without considering temporal effects:
\begin{itemize}
\item \emph{Imaging Quality} captures distortions such as blur, noise, and over-/under-exposure using the MUSIQ~\citep{ke2021musiq} image quality predictor.
\item \emph{Aesthetic Quality} reflects human-perceived aesthetic appeal using the LAION~\citep{laion_aesthetic_2022} aesthetic predictor.
\end{itemize}

\textbf{Temporal Quality} measures cross-frame consistency and motion:
\begin{itemize}
\item \emph{Motion Smoothness} measures whether motion is smooth by comparing interpolated frames from even-numbered frames using AMT~\citep{li2023amt} with the original odd-numbered frames.
\item \emph{Dynamic Degree} quantifies motion magnitude via optical-flow statistics using RAFT~\citep{teed2020raft}.
\item \emph{Subject Consistency} assesses whether subject appearance remains consistent over time using DINO~\citep{caron2021dino} features.
\item \emph{Background Consistency} evaluates temporal consistency of background scenes using CLIP~\citep{radford2021clip} image features.
\end{itemize}

\textbf{Semantics} measures text adherence:
\begin{itemize}
\item \emph{Scene Consistency} evaluates whether captions generated from the video~\citep{huang2023tag2text} match the intended scene described by the prompt.
\item \emph{Semantic\&Style Consistency} measures overall text--video alignment (including style consistency) computed by ViCLIP~\citep{wang2023internvid}. VBench refers to this metric as \emph{Overall Consistency}; we rename it to avoid the misleading implication that it is a representative aggregate over other metrics.
\end{itemize}

\subsection{Additional Metrics}
In addition to reporting the VBench dimension scores, we include the following widely used metrics. We use the 10-prompt and 48-prompt sets provided by Open-Sora (t2v\_sample and t2v\_sora, respectively), 93-prompt VBench \texttt{overall\_consistency.txt}, and the 101-prompt UCF-101~\citep{soomro2012ucf101} prompt set for these evaluations. As acknowledged in prior works~\citep{zhao2024vidit, feng2025q}, finite-sample FVD can yield unstable results; we therefore report FVD-FP16 across multiple prompt sets to improve reliability. All experiments are conducted with three random seeds.

\begin{itemize}
\item \emph{CLIPSIM} and \emph{CLIP-Temp} are computed using the EvalCrafter implementation~\citep{liu2024evalcrafter}: CLIPSIM averages frame-wise image--text similarity, while CLIP-Temp averages similarity between consecutive frames to reflect semantic temporal consistency.
\item \emph{DOVER VQA}~\citep{wu2023exploring} reports \emph{VQA-T} (technical quality; distortions such as noise/blur/exposure) and \emph{VQA-A} (aesthetic quality; composition, color harmony, naturalness).
\item \emph{Flow Score} uses RAFT~\citep{teed2020raft} to estimate dense optical flow between consecutive frames and averages the flow magnitude as a proxy for motion strength; following prior practice~\citep{feng2025q}, we report $\Delta$Flow (difference to the FP16 baseline) for robustness against pathological failures.
\item \emph{FVD-FP16} computes feature distribution distance between generated videos from FP16 model and quantized model using I3D features~\citep{blattmann2023align}; lower is better. It quantifies semantic and quality regression due to quantization. FVD-FP16 has a high correlation with human perception~\citep{zhao2024vidit}.
\end{itemize}

\section{Evaluation Setting Details}
\label{app:eval_setting}
Following prior work~\citep{zhao2024vidit, feng2025q}, for Open-Sora~1.0 we keep \texttt{embedder}, \texttt{t\_block}, and \texttt{final\_layer} in FP16. These modules lie at the network boundaries and account for a negligible fraction of the total computation. For Open-Sora~2.0, we similarly target only the linear layers in DiT blocks, \texttt{single\_blocks} and \texttt{double\_blocks}. We use FP16 weights and activations for the full-precision baseline. In our experiments, Open-Sora~1.0 generates 16 frames (8\,fps) at $512\times512$ resolution, while Open-Sora~2.0 generates 33 frames (24\,fps) at 256\,px resolution with a 16:9 aspect ratio. Weight dialect selection and quantization follow an MSE-based criterion.

We apply activation decomposition to quantization-sensitive layers: \texttt{attn\_temp.qkv} and \texttt{mlp.fc2} in Open-Sora~1.0, and \texttt{modulation.lin} and \texttt{txt\_mod.lin} in Open-Sora~2.0. For layers that precede attention score calculation (e.g., \texttt{attn\_temp.qkv}), salient tokens from the previous timestep are reused. We apply SeDA to MLP layers: \texttt{mlp.fc1} and \texttt{mlp.fc2} in Open-Sora~1.0, and \texttt{v\_proj}, \texttt{linear1} (fc1 equivalent), and \texttt{linear2} (fc2 equivalent) in \texttt{SingleStreamBlock} for Open-Sora~2.0. For \texttt{v\_proj} and \texttt{linear1}, which perform the first MLP operation yet precede attention in the same timestep, we reuse the 8-dialect sub-formatbook from the previous timestep. To mitigate SeDA overhead, we skip the first 20\% (40\%) of timesteps in Open-Sora~1.0 (2.0) and update anchor/correlated tokens every 10 (5) timesteps for block sizes 32 (16), with per-timestep updates in the last 10\% of timesteps.

In Open-Sora~1.0, \texttt{mlp.fc2} uses both activation decomposition and SeDA; we constrain both the original-value and residual quantization to use the sub-formatbook assigned by SeDA. For SeDA, we use attention thresholds of 8 (spatial) and 3 (temporal) for Open-Sora~1.0, with a $4\times4$ spatial tile and a $5\times5$ local correlation window. For Open-Sora~2.0, we use the same tile and window sizes and set the threshold to 5. For overlapping window regions, we assign them to the subsequent anchor token. 

All experiments are run on 80GB NVIDIA H100 GPUs (CUDA 12.6). Each video generation runs on a single GPU and takes tens to hundreds of seconds depending on the number format and model. Four GPUs are used in parallel to accelerate evaluation across prompts.

\section{Hardware Deployment Details}
\label{app:hardware}
\subsection{RTL Implementation} We modify two modules of the FlexASR~\citep{tambe202216} accelerator: \emph{PECore}, which handles main MAC computation and buffer management of processing elements (PEs), and \emph{GBControl}, which aggregates PE outputs and broadcasts data to PEs via control FSM logic and global shared buffer management. Specifically, we replace the input/output interface and MAC computation to support the target format in PECore, and add a quantization unit in GBControl. 

Each MAC has 8 lanes, where each lane handles block-wise (block size 16) multiply-accumulate operations per cycle. For FB4, each lane uses its own formatbook (Figure~\ref{fig:lut}a), shared among its multipliers. During synthesis, each index (0--7) is optimized separately (e.g., index 0 is mapped to constant 0, eliminating LUT retrieval). As a result, the total formatbook area accounts for only 12.6\% of the multiplier area (excluding accumulators). 

For the quantization unit, NVFP4 incurs higher latency and hardware complexity due to tensor-level scaling, which cannot be pipelined at block granularity, and floating-point block-wise scaling, which requires more cycles than the power-of-two scaling used by MXFP4 and FB4 (accomplished via exponent addition). Consequently, NVFP4 requires 6 cycles per block for quantization, while FB4 requires only one at 250MHz.

\subsection{CUDA Kernel Implementation}
We implement our CUDA kernel on H100 (CUDA 12.6), built on Open-Sora~2.0. Since \formatname uses power-of-two scaling and its normalized integer values (0--15) are exactly representable in FP8 e4m3, it is directly compatible with MXFP8 GEMM. However, H100 does not natively support MXFP8 GEMM, so we first attempt to use FP8 GEMM with fine-grained block-wise quantization. This incurs significant overhead, as H100 lacks native hardware support for fine-grained block-wise scaling required by MX formats. Instead, we use cuBLAS FP16 GEMM with on-the-fly weight and activation decoding from 4-bit to FP16, incorporating block-wise scale application. Quantization is fused with the preceding operation (e.g., GeLU, LayerNorm) where possible, except for flash attention kernels where fusion is not straightforward.

Peak memory usage is measured via PyTorch API. Since our implementation decodes to FP16 rather than FP8, we halve the decoded weight and activation sizes when estimating MXFP8-equivalent peak memory. Pre-decoded weight memory (stored in 4-bit) is unaffected.

Inference latency is measured via NVIDIA Nsight over multiple runs of the denoising stage and averaged. To conservatively estimate Blackwell MXFP8 latency, we: (1) profile the GEMM and decode wall-clock contribution excluding overlap with other operations; (2) halve GEMM time reflecting MXFP8's 2$\times$ throughput over FP16 on Blackwell; (3) halve decode time, as MXFP8 GEMM would at minimum halve memory read/write for decoding, and potentially eliminate it entirely with prologue fusion; and (4) halve the time where decode and GEMM overlap.

A large portion of the overhead stems from additional kernel launches introduced by \formatname quantization and SeDA, and limited prologue/epilogue fusion opportunities of cuBLAS GEMM on H100 preventing dequantization and quantization from being fused into the GEMM kernel. Therefore, we expect further optimizations such as CUDA graph adoption or CUTLASS-based MXFP8 GEMM kernels on the Blackwell platform to substantially reduce this overhead.

\section{Attention Scoring Design Choices}
\begin{table}[t]
\centering
\caption{Ablation of attention-score choices on Open-Sora~1.0.}
\label{tab:attn_score_ablation}
\fontsize{7.7}{8.8}\selectfont   
\setlength{\tabcolsep}{4pt}
\begin{tabular}{clcccccccc}
\toprule
\multicolumn{2}{c}{\shortstack[c]{\textbf{Method}\\\ }} &
\shortstack[c]{\textbf{Aesthetic}\\[-1pt]\textbf{Quality}} &
\shortstack[c]{\textbf{Imaging}\\[-2pt]\textbf{Quality}} &
\shortstack[c]{\textbf{Motion}\\[-1pt]\textbf{Smooth.}} &
\shortstack[c]{\textbf{Dynamic}\\[-2pt]\textbf{Degree}} &
\shortstack[c]{\textbf{Subject}\\[-2pt]\textbf{Consist.}} &
\shortstack[c]{\textbf{BG.}\\[-1pt]\textbf{Consist.}} &
\shortstack[c]{\textbf{Scene}\\[-1pt]\textbf{Consist.}} &
\shortstack[c]{\textbf{SS.}\\[-1pt]\textbf{Consist.}} \\
\midrule

(a) & \multicolumn{9}{l}{\textbf{Act. decomposition scoring}} \\
 & Incoming (avg. over queries) & 50.10 & 57.01 & 98.22 & 47.22 & 91.93 & 95.68 & 11.61 & 21.70 \\
 \rowcolor{gray!15}& Outgoing (avg. over keys)    & 51.25 & 60.00 & 97.62 & 48.61 & 90.74 & 95.04 & 17.04 & 21.99 \\

\midrule
(b) &\multicolumn{9}{l}{\textbf{SeDA anchor/corr. scoring}} \\
 & Post-softmax & 50.97 & 60.26 & 97.57 & 47.22 & 90.44 & 94.92 & 16.74 & 22.08 \\
\rowcolor{gray!15}
 & Pre-softmax  & 51.22 & 60.40 & 97.61 & 47.22 & 91.17 & 94.98 & 20.61 & 22.32 \\
\bottomrule
\end{tabular}
\end{table}

\label{app:attn_scoring}
\subsection{Incoming vs.\ outgoing attention} 
As shown in Table~\ref{tab:attn_score_ablation}a, we find that ranking tokens by the mean outgoing attention (i.e., selecting query tokens that strongly attend to other tokens on average) performs better than ranking by mean incoming attention (i.e., how much a token is attended by others). A plausible explanation is that highly attended tokens are often generic or globally shared tokens (e.g., background/context tokens), whereas tokens with strong outgoing attention are more likely to capture distinctive, semantically informative content. We also observe \emph{attention sinks} — tokens at frame corners and edges that attract attention from most other tokens — further making incoming attention an unfavorable criterion for anchor selection.

\subsection{Self-attention vs.\ cross attention}
Cross-attention in Open-Sora 1.0 injects text conditions into visual tokens, and in Open-Sora 2.0’s multimodal attention (with concatenated text and image tokens), cross-modal interactions occur similarly. Prior work on diffusion models suggests that cross-attention maps often provide useful correspondences between text tokens and visual regions, while self-attention tends to capture spatial structural patterns less directly tied to semantic categories~\citep{wen2025analysis,liu2024towards,cai2025ditctrl}.

However, our goal is not exact text-to-region grounding; rather, we need a reliable measure of token-level semantic relatedness. For this purpose, cross-attention is not ideal because text prompts contain many semantically weak tokens (e.g., a, the, of). Without an additional online stage to identify semantically informative text tokens, cross-attention may overemphasize weakly informative function or stylistic words.

Therefore, we use self-attention scores to estimate semantic relatedness for SeDA and for salient-token selection in activation decomposition. We leave leveraging cross-attention for more precise semantic extraction to future work.

\subsection{Pre-softmax vs.\ post-softmax scores}
We apply attention-score thresholding to filter out weakly correlated tokens. Empirically, pre-softmax thresholding performs better (Table~\ref{tab:attn_score_ablation}b) and is simpler to use for two reasons. First, post-softmax scores are inherently concentrated on a few large values, which can suppress moderately but meaningfully correlated tokens and make them harder to retain with thresholding. Second, this concentration makes threshold selection less robust, since the effective score range varies with the input distribution.

\begin{table}[t]
\caption{Ablation of salient token selection and attention scoring configurations for activation decomposition (w/o SeDA) on Open-Sora 1.0. ``cond'' denotes allocating the full salient token budget to the conditional branch only (rather than splitting it across conditional and unconditional branches).}
\label{tab:salient_token_scoring_ablation}
\centering
{\scriptsize
\setlength{\tabcolsep}{2.8pt} 
\renewcommand{\arraystretch}{1.06}
\begin{tabular}{ccccccccccc}
\toprule
\multicolumn{3}{c}{\shortstack[c]{\textbf{Method}\\\ }} &
\shortstack[c]{\textbf{Aesthetic}\\[-1pt]\textbf{Quality}} &
\shortstack[c]{\textbf{Imaging}\\[-2pt]\textbf{Quality}} &
\shortstack[c]{\textbf{Motion}\\[-1pt]\textbf{Smooth.}} &
\shortstack[c]{\textbf{Dynamic}\\[-2pt]\textbf{Degree}} &
\shortstack[c]{\textbf{Subject}\\[-2pt]\textbf{Consist.}} &
\shortstack[c]{\textbf{BG.}\\[-1pt]\textbf{Consist.}} &
\shortstack[c]{\textbf{Scene}\\[-1pt]\textbf{Consist.}} &
\shortstack[c]{\textbf{SS.}\\[-1pt]\textbf{Consist.}} \\
\cmidrule(lr){1-3}\cmidrule(lr){4-11}

\multicolumn{2}{l}{\textbf{Target selection}} & \textbf{Timestep range}
& & & & & & & & \\
\midrule

\multicolumn{2}{l}{Full tensor}                 & Steps 0--24  & 47.90 & 54.30 & 97.98 & 41.67 & 90.28 & 95.42 & 18.01 & 20.66 \\
\multicolumn{2}{l}{Full tensor}                 & Steps 75--99 & 46.19 & 49.68 & 98.57 & 43.06 & 90.74 & 95.41 & 11.46 & 20.12 \\
\multicolumn{2}{l}{Random (25\%)}               & All steps    & 51.14 & 58.23 & 98.10 & 47.22 & 91.15 & 95.74 & 16.96 & 21.46 \\
\multicolumn{2}{l}{Top-25\% magnitude}          & All steps    & 50.27 & 54.57 & 98.10 & 36.11 & 91.65 & 95.63 & 13.10 & 21.74 \\
\rowcolor{gray!15}
\multicolumn{2}{l}{Top-1 mean attn / \(1{\times}4\) tile} 
                                               & All steps    & 51.25 & 60.00 & 97.62 & 48.61 & 90.74 & 95.04 & 17.04 & 21.99 \\

\midrule

\textbf{Spatial scoring} & \textbf{Attn agg. tile} & \textbf{Temporal scoring}
& & & & & & & & \\
\midrule

Raw        & \(4{\times}4\)   & Raw         & 48.32 & 55.38 & 97.98 & 48.61 & 90.86 & 95.00 & 16.37 & 21.49 \\
ReLU       & \(4{\times}4\)   & ReLU        & 48.34 & 54.58 & 97.95 & 40.28 & 90.44 & 94.88 & 13.39 & 21.49 \\
ABS        & \(4{\times}4\)   & ABS         & 48.26 & 55.20 & 97.96 & 44.44 & 90.27 & 94.90 & 16.00 & 21.59 \\ 
ABS        & \(4{\times}4\)   & ReLU         & 47.71 & 55.08 & 97.98 & 45.83 & 90.75 & 94.91 & 16.59 & 21.52 \\ \hline
Raw        & \(4{\times}4\)   & Raw (cond)  &50.86 & 59.80 & 97.64 & 51.39 & 90.67 & 95.21 & 17.41 & 22.09\\
ABS        & \(4{\times}4\)   & Raw (cond)  &50.73 & 59.42 & 97.61 & 47.22 & 90.41 & 94.85 & 18.15 & 22.17\\
Raw        & \(4{\times}4\)   & ReLU (cond)  &50.79 & 59.89 & 97.61 & 50.00 & 90.50 & 94.83 & 15.40 & 22.09\\
ABS        & \(4{\times}4\)   & ABS (cond)  & 50.81 & 60.13 & 97.63 & 48.61 & 90.83 & 94.90 & 14.81 & 22.10 \\
\rowcolor{gray!15}
ABS        & \(4{\times}4\)   & ReLU (cond) & 51.25 & 60.00 & 97.62 & 48.61 & 90.74 & 95.04 & 17.04 & 21.99 \\
ABS (cond)  & \(4{\times}4\)   & ReLU (cond) & 46.50 & 47.74 & 97.16 & 40.28 & 92.99 & 96.51 & 6.92 & 17.04 \\ \hline \rule{0pt}{2.4ex}
ABS        & \(32{\times}32\) & ReLU (cond) & 51.30&	59.41&	97.68&	48.61&	90.87&	95.19&	16.82&	22.14 \\

\bottomrule
\end{tabular}
}
\end{table}

\section{Activation Decomposition Salient Token Selection Criteria}
\label{app:salient_token}
Table~\ref{tab:salient_token_scoring_ablation} compares salient token selection strategies for activation decomposition. 

\para{Comparison with Alternative Selection Strategies}
Full-tensor decomposition, whether applied to early or late timestep ranges, underperforms attention-guided selection across most metrics. Random (25\%) and top-25\%-magnitude selection provide more balanced results but still fall short, particularly in Imaging Quality and Scene Consistency, confirming the effectiveness of attention-based token scoring. 

\para{Token Scoring} 
We compare raw attention scores with ABS/ReLU-transformed scores. Without conditional-branch awareness, Raw-Raw, ABS-ReLU, and ABS-ABS achieve similar quality. Once conditional-branch awareness is applied, Raw-Raw (cond) and ABS-ReLU (cond) remain comparable. We adopt ABS-ReLU as it scores higher on frame-wise quality metrics (Aesthetic and Imaging Quality), which are particularly important given that SeDA forces a non-MSE-optimal format and may affect per-frame fidelity. Also, ABS/ReLU transformations prevent large negative scores from canceling meaningful positive correlations, providing a more stable scoring signal.

\para{Conditional Branch Awareness}
With a fixed salient-token budget, we compare splitting the budget across conditional/unconditional branches versus allocating the full budget to the conditional branch. For \texttt{attn\_temp.qkv}, which relies on temporal attention scoring, conditional-only selection performs substantially better; however, applying the same strategy to \texttt{mlp.fc2}, which uses spatial attention scoring, severely degrades quality (e.g., scene consistency drops to 6.92). We attribute this to classifier-free guidance, where the update depends on the conditional--unconditional activation delta: improving only the conditional branch can strengthen the conditioning signal, but a noisier unconditional branch can corrupt the delta. This effect is stronger for \texttt{mlp.fc2}, which directly produces output activations, whereas \texttt{attn\_temp.qkv} is followed by attention/normalization/linear layers that can partially absorb unconditional-branch noise.

\para{Attention Score Aggregation Tile Size}
Finally, selecting salient tokens within each $4\times4$ tile encourages uniform spatial coverage across the frame. By contrast, global frame-level selection often concentrates tokens in a few regions---typically large background areas that contain many tokens---leading to poor coverage elsewhere and requiring much higher computational cost while achieving comparable quality.

\section{SeDA Ablation Studies}
\label{app:seda_ablation}
\begin{table}[t]
\centering
\caption{SeDA configuration ablations on Open-Sora~1.0. In the timestep update schedule, the update period denotes the token-update interval in timesteps (1 = every timestep); ``$x$ (20--89)'' means period $x$ on timesteps 20--89 and period 1 otherwise.}
\label{tab:seda_ablation}
\fontsize{7.7}{8.8}\selectfont
\setlength{\tabcolsep}{3.2pt}
\begin{tabular}{c ccccccccccc}
\toprule
\multicolumn{3}{c}{\shortstack[c]{\textbf{Method}\\\ }} &
\shortstack[c]{\textbf{Aesthetic}\\[-1pt]\textbf{Quality}} &
\shortstack[c]{\textbf{Imaging}\\[-2pt]\textbf{Quality}} &
\shortstack[c]{\textbf{Motion}\\[-2pt]\textbf{Smooth.}} &
\shortstack[c]{\textbf{Dynamic}\\[-2pt]\textbf{Degree}} &
\shortstack[c]{\textbf{Subject}\\[-2pt]\textbf{Consist.}} &
\shortstack[c]{\textbf{BG.}\\[-1pt]\textbf{Consist.}} &
\shortstack[c]{\textbf{Scene}\\[-1pt]\textbf{Consist.}} &
\shortstack[c]{\textbf{SS.}\\[-1pt]\textbf{Consist.}} \\
\midrule

(a)&\multicolumn{2}{c}{\textbf{Temporal-axis integration}} \\
\midrule

& \multicolumn{2}{c}{Per-frame anchors (2D SeDA)} 
& 51.49 & 60.93 & 97.55 & 45.83 & 91.04 & 94.74 & 14.73 & 22.17 \\
\rowcolor{gray!15}
& \multicolumn{2}{c}{Cross-frame main anchor (3D SeDA)}
& 51.22 & 60.40 & 97.61 & 47.22 & 91.17 & 94.98 & 20.61 & 22.32 \\

\midrule
(b) & \textbf{Anchor tile} & \textbf{Correlation window}
& & & & & & & & \\
\midrule

& $4\times4$ & $7\times7$
& 50.88 & 60.11 & 97.59 & 48.61 & 91.15 & 95.00 & 18.90 & 22.17 \\
& $4\times4$ & $3\times3$
& 51.07 & 60.18 & 97.58 & 43.06 & 90.67 & 94.81 & 19.12 & 21.92 \\
\rowcolor{gray!15}
& $4\times4$ & $5\times5$
& 51.22 & 60.40 & 97.61 & 47.22 & 91.17 & 94.98 & 20.61 & 22.32 \\
& $8\times8$ & $7\times7$
& 51.19 & 60.65 & 97.56 & 52.78 & 90.99 & 94.86 & 18.75 & 22.28 \\
& $2\times2$ & $3\times3$
& 50.93 & 60.32 & 97.63 & 51.39 & 91.05 & 94.82 & 15.63 & 22.17 \\

\midrule
(c) & \textbf{Timesteps} & \textbf{Update period}
& & & & & & & & \\
\midrule

 & 0--99 & 1
& 51.48 & 60.54 & 97.58 & 50.00 & 90.92 & 95.04 & 16.74 & 22.32 \\
& 20--99 & 1
& 51.45 & 60.99 & 97.58 & 45.83 & 91.14 & 95.04 & 16.15 & 21.99 \\
& 20--99 & 10
& 50.75 & 60.46 & 97.60 & 41.67 & 90.74 & 94.96 & 15.48 & 22.16 \\
\rowcolor{gray!15}
& 20--99 & 10 (20--89)
& 51.22 & 60.40 & 97.61 & 47.22 & 91.17 & 94.98 & 20.61 & 22.32 \\
& 20--99 & 5 (20--89)
& 51.41 & 60.43 & 97.59 & 50.00 & 91.09 & 94.94 & 18.30 & 22.18 \\
& 20--99 & 20 (20--89)
& 51.10 & 60.58 & 97.60 & 47.22 & 91.02 & 94.98 & 15.40 & 22.11 \\
\bottomrule
\end{tabular}
\end{table}

\para{Temporal Axis Integration}
Because Open-Sora~1.0 uses factorized attention, we first identify anchor candidates for each frame and then select a single anchor token with the highest temporal mean attention score across frames. In Table~\ref{tab:seda_ablation}a, we compare a \emph{per-frame} SeDA strategy (applying SeDA independently to each frame-level anchor and its correlated tokens) with our \emph{temporal-axis integration} strategy (selecting a single anchor across frames for the same spatial tile and excluding frames with low attention scores from that anchor). While per-frame SeDA shows advantages in frame-wise quality (Aesthetic, Imaging Quality), temporal-axis integration outperforms it on other metrics, with the most notable improvement in scene consistency, likely because cross-frame anchors better capture scene-level semantic content shared across frames.

\para{Anchor Tile and Local Correlation Window Size}
Table~\ref{tab:seda_ablation}b compares different anchor tile and local correlation window sizes in Open-Sora 1.0. A smaller local correlation window tends to limit SeDA's effectiveness by reducing the number of correlated tokens. A larger window can improve some metrics but forces the same sub-formatbook on too many tokens and increases overlap between neighboring windows, making accurate anchor--correlated token pairing more difficult. Similarly, overly small or large anchor tile sizes also degrade performance, suggesting that both parameters benefit from careful balancing.

\para{Timesteps for Applying SeDA Under Overhead Constraints}
Table~\ref{tab:seda_ablation}c compares when SeDA is applied and how often anchor/correlated tokens are updated over denoising timesteps. Skipping SeDA during the initial \(\sim\)20\% of denoising causes minimal degradation, likely because early-stage attention maps are unstable and activations remain noise-dominated, making anchor/correlated token selection unreliable. Updating tokens every 10 timesteps in the intermediate region has little effect on performance, whereas infrequent updates in the final 10\% of timesteps degrade quality, consistent with the non-negligible final-step anchor-token variation observed in the profiling results (Figure~\ref{fig:seda}c). Results with 5- and 20-timestep update intervals further confirm that a balanced update frequency is important to preserve both trackability and consistency.

\section{Comparison to \formatname and NVFP4}
\label{app:nvfp4}
\begin{table}[t]
\centering
\caption{Comparison between NVFP4 and \formatname (w/o activation decomposition, SeDA) on additional metrics on Open-Sora 1.0. \emph{num\_group} is 8 for \formatname.}
\label{tab:nvfp4_vs_semantic_additional}
\setlength{\tabcolsep}{4pt}
\fontsize{7.7}{8.8}\selectfont
\renewcommand{\arraystretch}{1.15}
\begin{tabular}{c c c c c c c c c}
\toprule
\shortstack[c]{\textbf{Prompt} } &
\shortstack[c]{\textbf{Method} } &
\shortstack[c]{\textbf{Block size}} &
\shortstack[c]{\textbf{FVD-FP16} ($\downarrow$) } &
\shortstack[c]{\textbf{CLIP-Temp} } &
\shortstack[c]{\textbf{CLIPSIM}} &
\shortstack[c]{\textbf{VQA-A}} &
\shortstack[c]{\textbf{VQA-T}} &
\shortstack[c]{\textbf{$\Delta$ Flow}($\downarrow$)} \\
\hline
\multirow{4}{*}{t2v\_sample}
& NVFP4 & 16 & \textbf{1.12}  & 0.9978 & 0.1752 & 50.58 & \textbf{47.97} & 0.48 \\
& \formatname   & 16 & 1.24 & \textbf{0.9983} & \textbf{0.1781} & \textbf{50.72} & 47.87 & \textbf{0.26} \\ \cline{2-9}
& NVFP4 & 32 &  2.11&  0.9976&  0.1775&  49.91&  47.43&\textbf{0.21}  \\
& \formatname   & 32 &  \textbf{1.59}& \textbf{0.9986} & \textbf{0.1778} & \textbf{50.29} & \textbf{47.78} &  0.84\\
\hline
\multirow{4}{*}{UCF-101}
& NVFP4 & 16 & 67.08 & 0.9971 & 0.2021 & 49.69 & 47.31 & 3.49
 \\
& \formatname   & 16 & \textbf{61.80} & \textbf{0.9976} & \textbf{0.2072} & \textbf{49.89} & \textbf{47.54} & \textbf{1.33} \\ \cline{2-9}
& NVFP4 & 32 & \textbf{81.72} & \textbf{0.9973} & 0.2052 & \textbf{48.80} & 46.93 & \textbf{2.83} \\
& \formatname   & 32 & 85.86 & 0.9972 & \textbf{0.2061} & 48.77 & \textbf{46.93} & 3.91 \\
\bottomrule
\end{tabular}
\end{table}
We compare \formatname against NVFP4 on additional metrics using the t2v\_sample and UCF-101 prompt sets in Table~\ref{tab:nvfp4_vs_semantic_additional}. Both are evaluated with the same block size and without activation decomposition or SeDA to ensure a fair comparison at the same effective bit-width. Although a block size of 32 on the UCF-101 dataset is less favorable for \formatname, it outperforms NVFP4 in most cases.

\section{Additional Cumulative Impact Analysis}
\label{app:cumulative}
\begin{table}[t]
\centering
\caption{Stepwise quality improvement across multiple prompt sets (block size 32).}
\label{tab:ucf}
\setlength{\tabcolsep}{4pt}
\fontsize{7.7}{8.8}\selectfont
\renewcommand{\arraystretch}{1.15}
\begin{tabular}{l l c c c c c cc}
\toprule
\shortstack[c]{\textbf{Dataset}} &
\shortstack[c]{\textbf{Method} } &
\shortstack[c]{\textbf{FVD-FP16} ($\downarrow$) } &
\shortstack[c]{\textbf{CLIP-Temp} } &
\shortstack[c]{\textbf{CLIPSIM}} &
\shortstack[c]{\textbf{VQA-A}} &
\shortstack[c]{\textbf{VQA-T}} &
\shortstack[c]{\textbf{$\Delta$ Flow}($\downarrow$)} \\
\hline
& FP16      & -     & 0.9973 & 0.2044 & 51.11 & 47.96 & -    \\ \cline{2-8}
\multirow{3}{*}{UCF-101} & \formatname       & 85.86 & \textbf{0.9972} & \textbf{0.2061} & 48.77 & 46.93 & 3.91 \\
& +Decomp.  & \textbf{67.04} & 0.9972 & 0.2048 & 49.74 & 47.45 & 2.50 \\
& +SeDA     & 67.46 & 0.9971 & 0.2050 & \textbf{49.75} & \textbf{47.46} & \textbf{1.36}\\
\hline
& FP16      & -     & 0.9984 & 0.1953 & 51.54 & 48.15 & -    \\ \cline{2-8}
\multirow{3}{*}{VBench} & \formatname       & 62.96 & 0.9980 & \textbf{0.1916} & 49.48 & 47.31 & 2.68 \\
& +Decomp.  & 46.61 & \textbf{0.9982} & 0.1897 & 50.43 & 47.74 & 1.25 \\
& +SeDA     & \textbf{44.55} & 0.9981 & 0.1895 & \textbf{50.47} & \textbf{47.77} & \textbf{1.00}\\
\hline
& FP16      & -     & 0.9983 & 0.1955 & 51.45 & 48.16 & -    \\ \cline{2-8}
\multirow{3}{*}{t2v\_sora} & \formatname       & 27.31 & 0.9980 & 0.1869 & 49.37 & 47.22 & 4.99 \\
& +Decomp.  & 16.60 & \textbf{0.9983} & 0.1879 & 50.31 & 47.71 & 3.80 \\
& +SeDA     & \textbf{16.13} & 0.9983 & \textbf{0.1884} & \textbf{50.32} & \textbf{47.75} & \textbf{2.93}\\
\bottomrule
\end{tabular}
\end{table}
Table~\ref{tab:ucf} presents the cumulative impact of the three proposed techniques on the UCF-101, VBench (\texttt{overall\_consistency.txt}, 93 prompts), and t2v\_sora (48 prompts) prompt sets. FVD-FP16 and flow score generally improve as additional components are incorporated, indicating closer alignment with the FP16 baseline, with activation decomposition contributing the most substantial gains. 

\section{Combination with Another PTQ Method}
\label{app:rotation}
\begin{table}[t]
\centering
\caption{Synergy of rotation-based PTQ methods on Open-Sora 1.0.}
\label{tab:rotation}
\fontsize{7.7}{8.8}\selectfont   
\setlength{\tabcolsep}{3.5pt}
\renewcommand{\arraystretch}{1.1}

\begin{tabular}{ccccccccccc}

\toprule
\shortstack[c]{\textbf{Method}\\\ } &
\shortstack[c]{\textbf{Block}\\[-1pt]\textbf{Size}} &

\shortstack[c]{\textbf{Aesthetic}\\[-1pt]\textbf{Quality}} &
\shortstack[c]{\textbf{Imaging}\\[-2pt]\textbf{Quality}} &
\shortstack[c]{\textbf{Motion}\\[-1pt]\textbf{Smooth.}} &
\shortstack[c]{\textbf{Dynamic}\\[-2pt]\textbf{Degree}} &
\shortstack[c]{\textbf{Subject}\\[-2pt]\textbf{Consist.}} &
\shortstack[c]{\textbf{BG.}\\[-1pt]\textbf{Consist.}} &
\shortstack[c]{\textbf{Scene}\\[-1pt]\textbf{Consist.}} &
\shortstack[c]{\textbf{SS.}\\[-1pt]\textbf{Consist.}} \\
\hline
\papername & \multirow{2}{*}{16} & 54.49 & 63.72 & 97.66 & 34.72 & 93.28 & 95.86 & 22.92 & 23.47 \\
\papername+Rotation  &                    & 55.52 & 64.95 & 97.40 & 36.11 & 93.18 & 95.51 & 26.64 & 23.93 \\
\hline
\papername & \multirow{2}{*}{32} & 51.22 & 60.40 & 97.61 & 47.22 & 91.17 & 94.98 & 20.61 & 22.32 \\
\papername+Rotation  &                    & 55.07 & 62.70 & 97.31 & 40.28 & 92.82 & 95.65 & 24.40 & 23.74 \\
\hline
\end{tabular}

\end{table}

Table~\ref{tab:rotation} demonstrates additional gains when \papername is combined with Hadamard rotation~\citep{ashkboos2024quarot}. Notably, aesthetic quality, imaging quality, and scene consistency improve significantly. Rotation spreads outliers across the tensor, reducing per-block outliers and making value distributions more approximately Gaussian. We hypothesize that while finding an optimal dialect for heavily skewed distributions with outliers is challenging, the less skewed post-rotation distributions are easier to capture with our formatbook, amplifying the benefit of mixed-format quantization. We exclude rotation from \papername's default configuration to isolate the contribution of our proposed techniques, given that rotation is an independently established PTQ method.

\section{Broader Impacts}
\label{app:impact}
\papername improves the accessibility of local video generation on edge devices using 4-bit quantization. However, as with all video generation methods, it could potentially lower the barrier to misuse for generating deepfakes or disinformation. Nonetheless, \papername is a post-training quantization method that does not introduce new generative capabilities beyond those of the base models.
\newpage
\section{More Qualitative Results}
\begin{figure}[H]
  \centering
  \includegraphics[page=6, width=0.9\columnwidth]{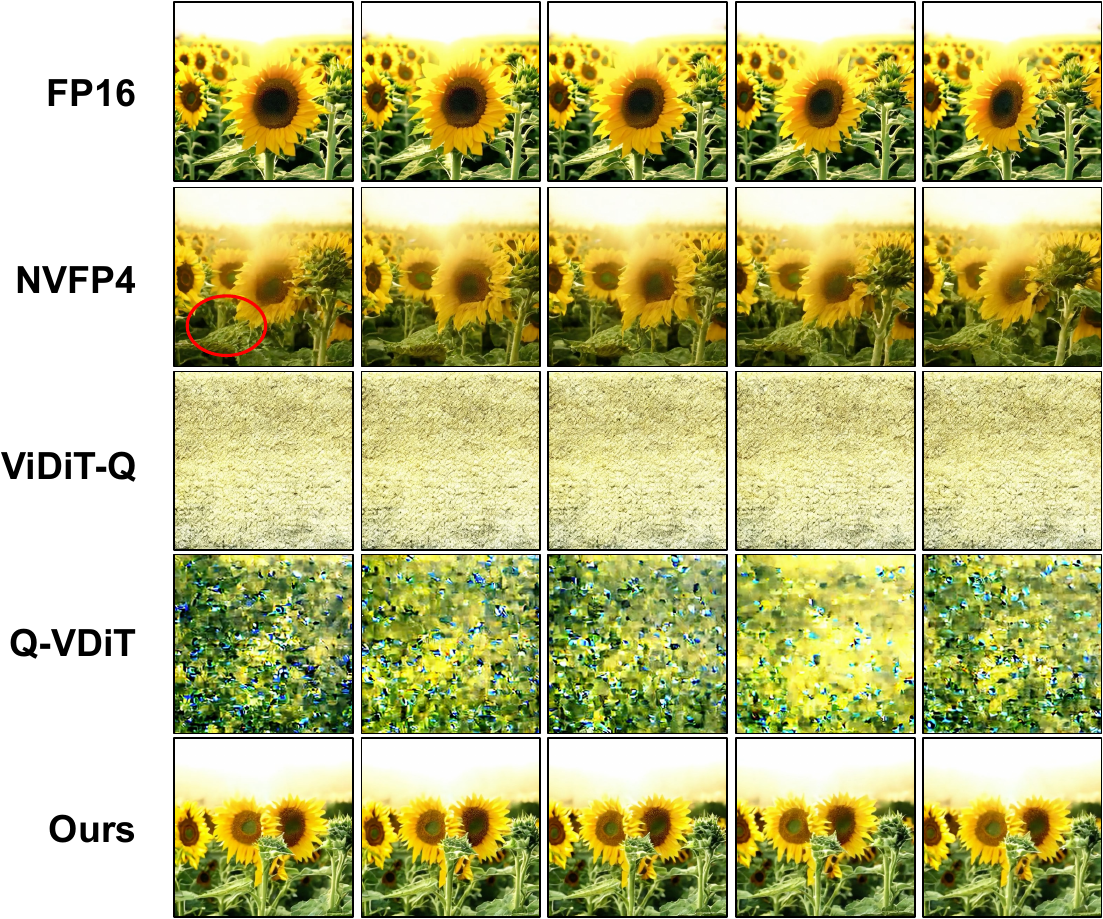}
\caption{Qualitative results on Open-Sora~1.0 for the prompt: “The vibrant beauty of a sunflower field \ldots”. Effective bit-width (A/W): NVFP4 = 4.5/4.5, Ours = 4.76/4.31. NVFP4 shows noisy leaves whose shapes are difficult to recognize throughout the video and exhibits poor structural consistency. In contrast, \papername preserves the sunflower and leaf structures more clearly and maintains better visual consistency over time.}
  \label{fig:qual1}
\end{figure}

\begin{figure}[t]
  \centering
  \includegraphics[page=7, width=0.9\columnwidth]{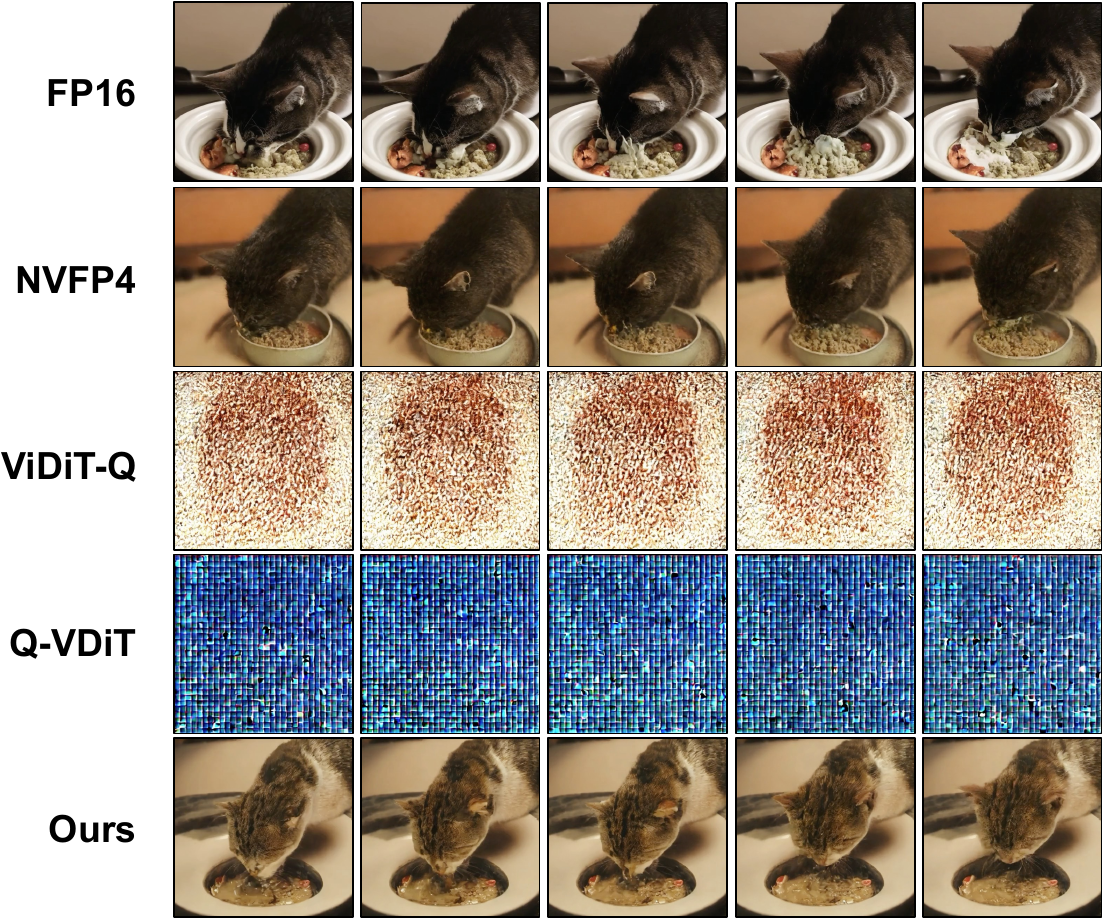}
\caption{Qualitative results on Open-Sora~1.0 for the prompt: “A cat eating food out of a bowl”. Effective bit-width (A/W): NVFP4 = 4.5/4.5, Ours = 4.76/4.31. NVFP4 frequently produces frames where the cat and food become entangled, 
with the food in the bowl exhibiting significant noise and flickering 
artifacts. FP16 also occasionally produces such entangled frames but remains 
visually sharp overall. In contrast, \papername maintains more stable cat and 
food shapes compared to NVFP4 and achieves better imaging quality.}
  \label{fig:qual2}
\end{figure}

\begin{figure}[t]
  \centering
  \includegraphics[page=8, width=0.9\columnwidth]{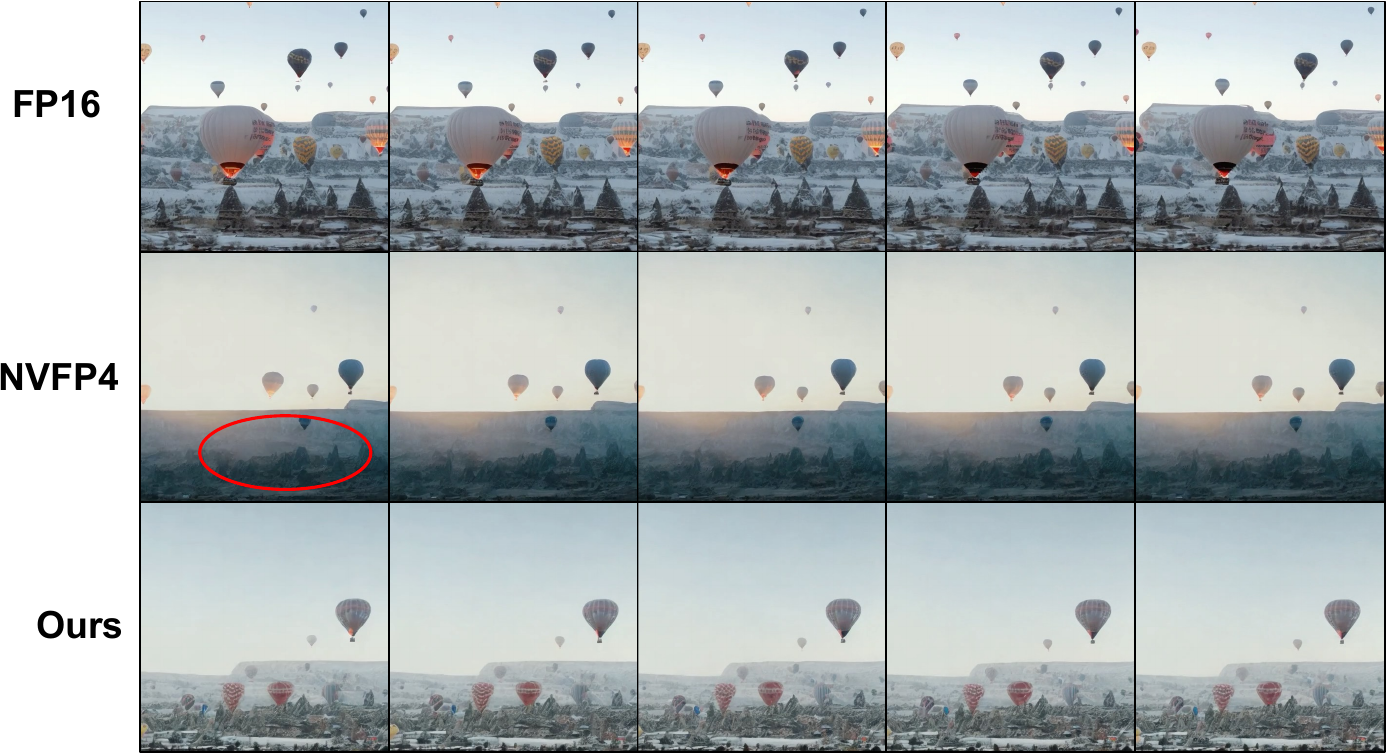}
\caption{Qualitative results on Open-Sora~1.0 for the prompt: “A vibrant scene of a snowy mountain landscape \ldots”. Effective bit-width (A/W): NVFP4 = 4.5/4.5, Ours = 4.76/4.31. NVFP4 shows a blurred background due to lower imaging quality. In contrast, \papername preserves clearer background details and more detailed hot-air balloons, resulting in a more visually faithful scene.}
  \label{fig:qual3}
\end{figure}

\begin{figure}[t]
  \centering
  \includegraphics[page=10, width=0.9\columnwidth]{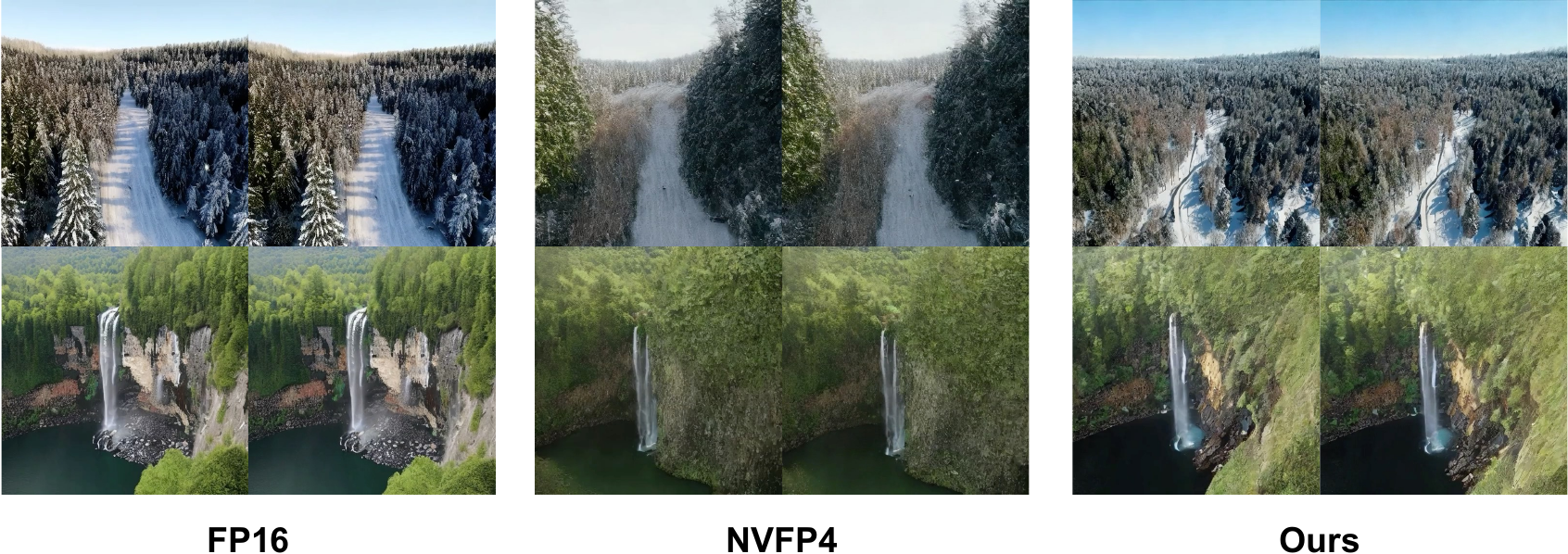}
\caption{Qualitative results on Open-Sora~1.0 for the prompts “A snowy forest landscape with a dirt road \ldots” and “The video captures the majestic beauty of a waterfall \ldots”. Effective bit-width (A/W): NVFP4 = 4.5/4.5, Ours = 4.76/4.31. \papername produces higher-quality videos that more closely match FP16 than NVFP4.}

  \label{fig:qual4}
\end{figure}

\begin{figure}[t]
  \centering
  \includegraphics[page=9, width=0.9\columnwidth]{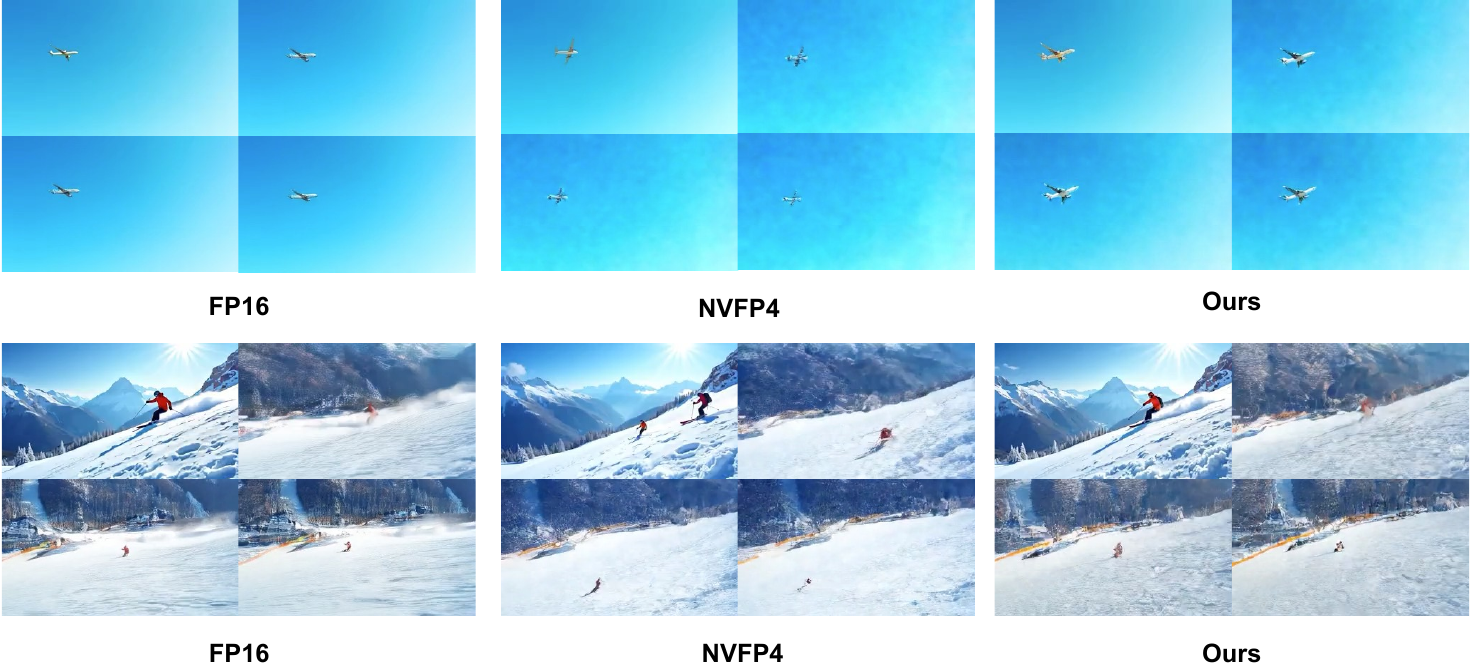}
\caption{Qualitative results on Open-Sora~2.0 for the prompt: “an airplane soaring through a clear blue sky” and "ski slope". Effective bit-width (A/W): NVFP4 = 4.25/4.25, Ours = 4.31/4.31. \papername shows better subject quality over NVFP4.}
  \label{fig:qual5}
\end{figure}

\begin{figure}[t]
  \centering
  \includegraphics[page=12, width=0.9\columnwidth]{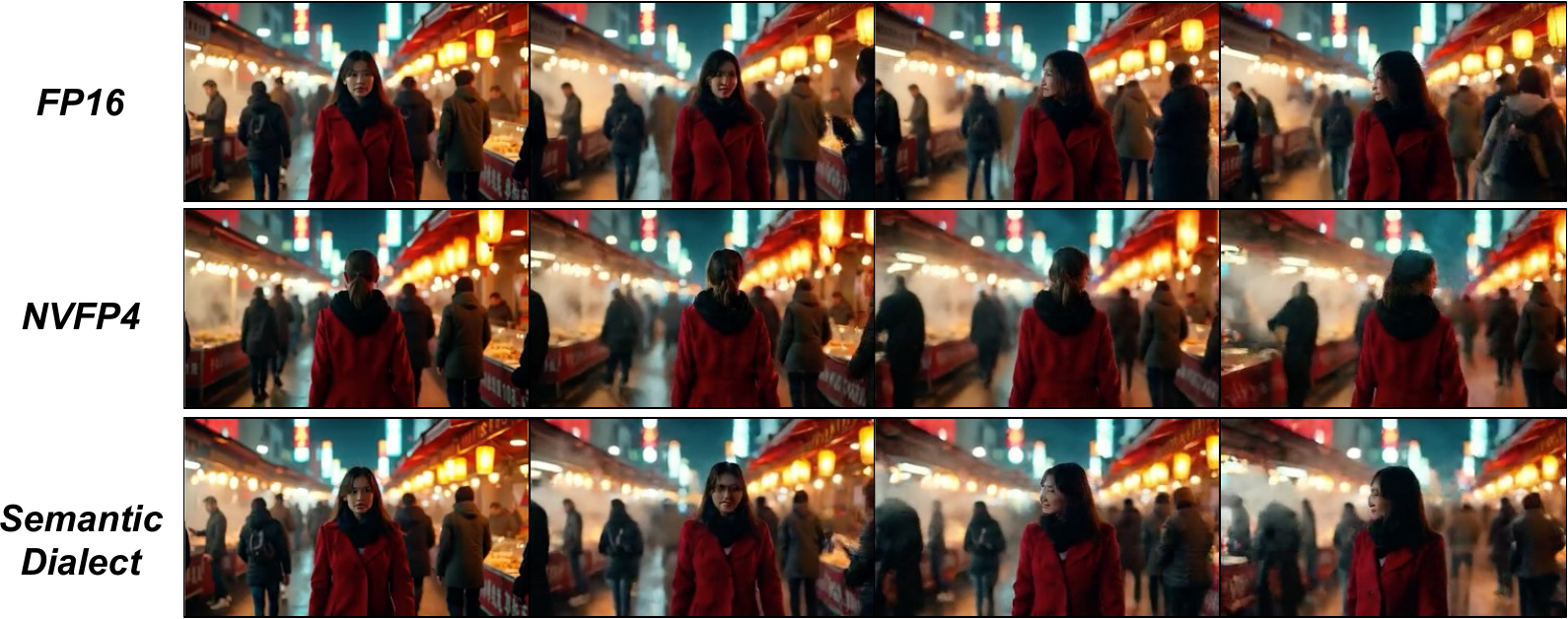}
\caption{Qualitative results on Open-Sora~2.0 for the prompt: “A woman in a red coat and black scarf walks through a crowded night market\ldots". Effective bit-width (A/W): NVFP4 = 4.5/4.5, Ours = 4.63/4.63.}
  \label{fig:qual6}
\end{figure}

\begin{figure}[t]
  \centering
  \includegraphics[page=13, width=0.9\columnwidth]{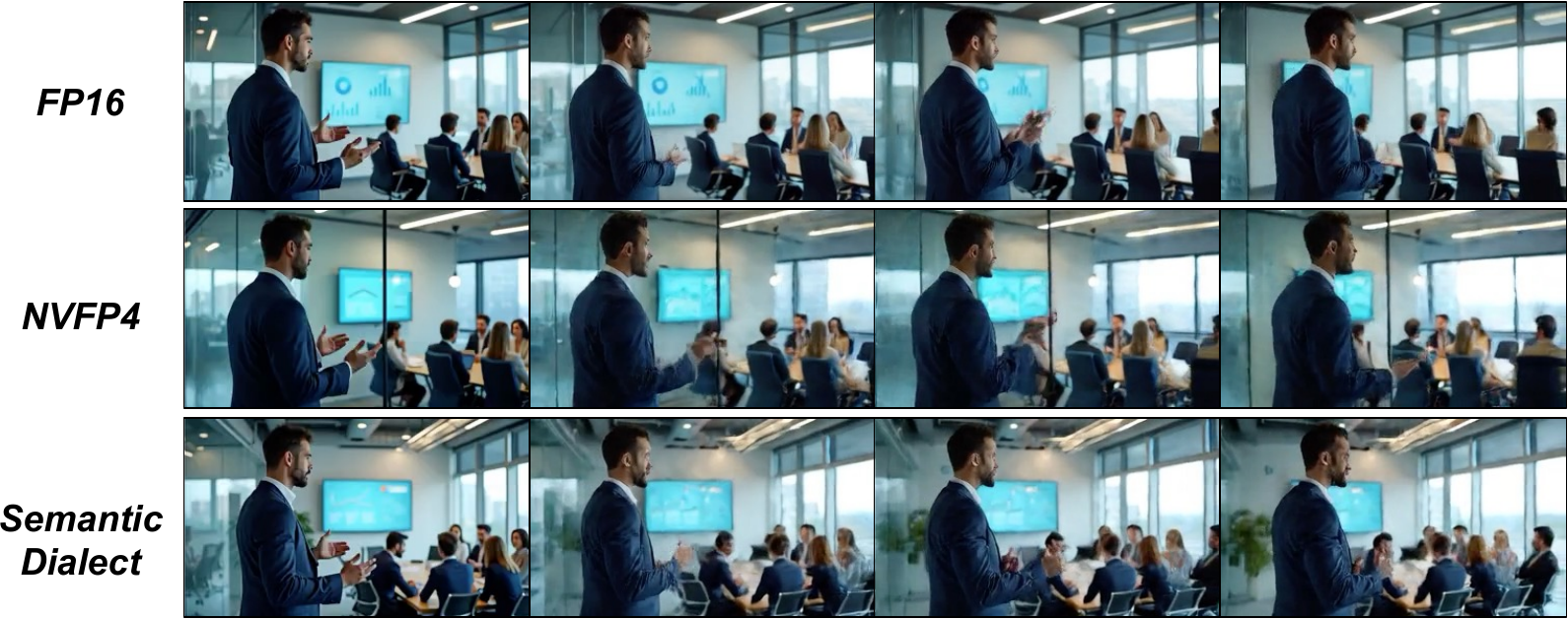}
\caption{Qualitative results on Open-Sora~2.0 for the prompt: “A man in a navy suit gives a presentation in a modern glass-walled office meeting room\ldots". Effective bit-width (A/W): NVFP4 = 4.5/4.5, Ours = 4.63/4.63.}
  \label{fig:qual7}
\end{figure}

\begin{figure}[t]
  \centering
  \includegraphics[page=14, width=0.9\columnwidth]{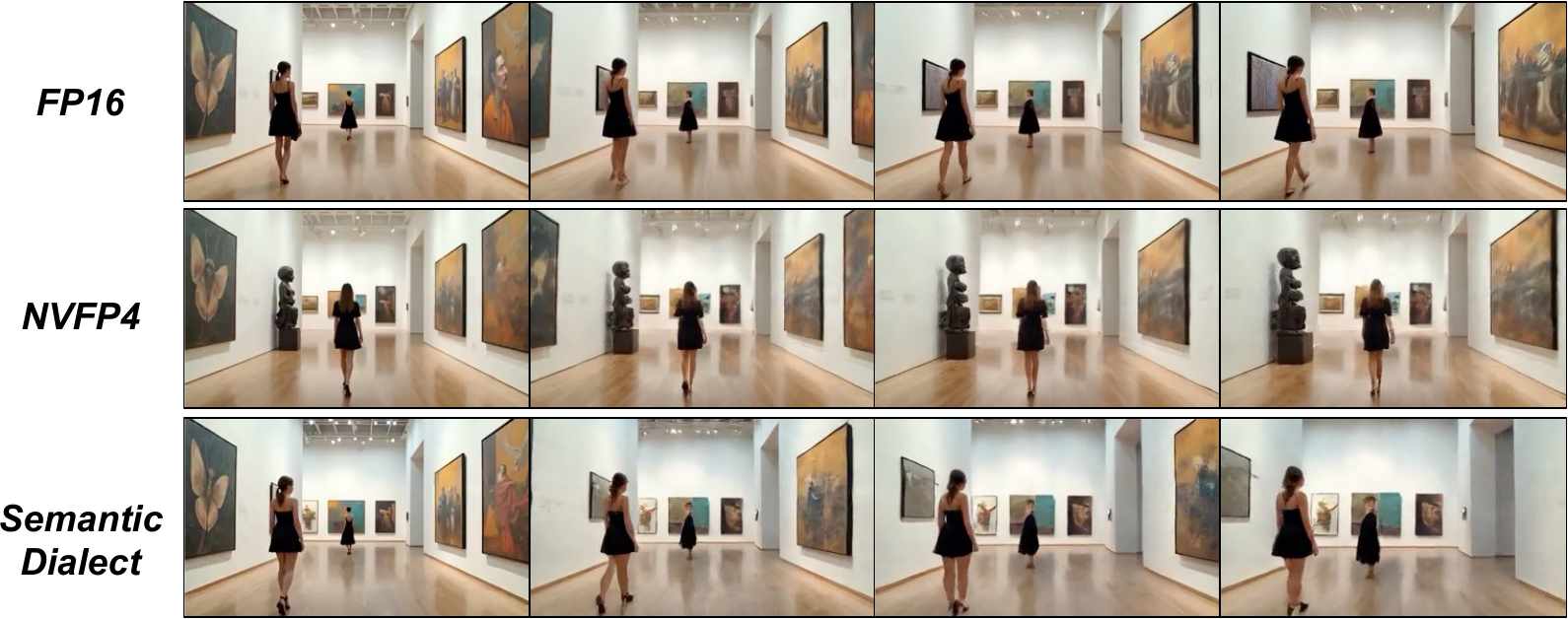}
\caption{Qualitative results on Open-Sora~2.0 for the prompt: “A woman in a black dress slowly walks through a contemporary art museum gallery with large paintings\ldots". Effective bit-width (A/W): NVFP4 = 4.5/4.5, Ours = 4.63/4.63.}
  \label{fig:qual8}
\end{figure}

\end{document}